\documentclass[letterpaper]{article} 
\usepackage[draft]{aaai2026}  
\usepackage{times}  
\usepackage{helvet}  
\usepackage{courier}  
\usepackage[hyphens]{url}  
\usepackage{graphicx} 
\usepackage{natbib}  
\usepackage{caption} 
\usepackage{algorithm}
\usepackage{algorithmic}

\usepackage{newfloat}
\usepackage{listings}
\DeclareCaptionStyle{ruled}{labelfont=normalfont,labelsep=colon,strut=off} 
\floatstyle{ruled}
\newfloat{listing}{tb}{lst}{}
\floatname{listing}{Listing}
\usepackage{amsmath}
\usepackage{amsfonts}
\usepackage{booktabs}
\usepackage[dvipsnames]{xcolor}
\definecolor{best}{HTML}{D0EAF8}    
\definecolor{worst}{HTML}{F8D0D0}   
\usepackage[framemethod=TikZ]{mdframed}
\usepackage{multirow}
\usepackage{makecell}
\usepackage{enumitem}
\usepackage{colortbl}
\usepackage[most]{tcolorbox}
\usepackage{subcaption}

\title{Timing the Message: Language-Based Notifications for Time-Critical\\ Assistive Settings}
\author{
    Ya-Chuan Hsu $^{\rm 1,2,}$\footnotemark[2],
    Jonathan DeCastro \textsuperscript{\rm 2},
    Andrew Silva \textsuperscript{\rm 2},
    Guy Rosman \textsuperscript{\rm 2}
}
\affiliations{
    \textsuperscript{\rm 1}University of Southern California\\
    \textsuperscript{\rm 2}Toyota Research Institute\\
}

\begin{document}
\maketitle

\footnotetext[2]{Work conducted while Ya-Chuan Hsu was an intern at Toyota Research Institute.}
\begin{abstract}
    In time-critical settings such as assistive driving, assistants often rely on alerts or haptic signals to prompt rapid human attention, but these cues usually leave humans to interpret situations and decide responses independently, introducing potential delays or ambiguity in meaning. Language-based assistive systems can instead provide instructions backed by context, offering more informative guidance. However, current approaches (e.g., social assistive robots) largely prioritize content generation while overlooking critical timing factors such as verbal conveyance duration, human comprehension delays, and subsequent follow-through duration. These timing considerations are crucial in time-critical settings, where even minor delays can substantially affect outcomes. We aim to study this inherent trade-off between timeliness and informativeness by framing the challenge as a sequential decision-making problem using an augmented-state Markov Decision Process. We design a framework combining reinforcement learning and a generated offline taxonomy dataset, where we balance the trade-off while enabling a scalable taxonomy dataset generation pipeline. Empirical evaluation with synthetic humans shows our framework improves success rates by over 40\% compared to methods that ignore time delays, while effectively balancing timeliness and informativeness. It also exposes an often-overlooked trade-off between these two factors, opening new directions for optimizing communication in time-critical human-AI assistance.

\end{abstract}


\section{Introduction}
Language is commonly used for coordination in embodied assistive tasks (e.g., human-human, human-robot interactions), facilitating intent communication, establishing mutual knowledge, and providing actionable instructions~\cite{nikolaidis2018planning,hoffman2004collaboration,st2015robot}. 
In this work, we focus specifically on time-critical scenarios, which are prevalent in daily life yet remain underexplored in language-based assistive systems. 
More particularly, we are interested in the inherent trade-off between language-based notification timeliness and informativeness. 
Consider a common scenario: during a highway merge, a driving assistant saying ``Ease off the gas, there's a car entering from your right.'' provides necessary clarity but may take too long to convey and comprehend to react for collision avoidance, whereas ``Slow down!'' prompts rapid but potentially hazardous overreactions and result in rear-ending situations due to lack of context. 
Thus, effective language-based notifications in such scenarios must balance immediacy with sufficient informativeness to guide immediate and subsequent actions. 
This paper addresses the critical trade-off between timeliness and informativeness in the context of designing a \textit{language-based assistive notifier agent} for \textit{human-AI collaboration} in \textit{time-critical} contexts.

Prior works often emphasize providing notifications that are either timely (e.g., safety-critical alerts)~\cite{sundareswara2013using, li2024optimal, zhao2021efficient} or informative (e.g., teacher-student interactions)~\cite{reimann2024survey, shvo2023proactive} independently, without explicitly addressing their trade-off. While recent works have begun recognizing that informative notifications incur time costs~\cite{liu2024llm, zhang2023building}, the trade-off has yet to be formalized in a sequential decision-making context.

Motivated by this gap, we investigate the problem setting where
the human executes a task based on their existing knowledge, the assistive agent (i.e., the \textit{notifier}) solves a sequential decision-making task through (1) detecting knowledge gaps through human behavior, (2) determining the missing information, and (3) delivering the notification at a length that the human can comprehend before failure occurs. 
Additionally, while focusing on the time-critical settings, we select notifications that provide actionable instructions while including context when given sufficient time. 

To define the action space of our framework, we model each language notification as an action parameterized by three domain-agnostic properties: topic, the word-by-word point of incremental comprehension~\cite{chromik2017incremental}, and total length. To gather a dataset of notifications at scale would be impractical. For instance, a 10-word utterance would require 
hundreds of annotations.

Recent advances show that Large Language Models (LLMs) approximate human-like comprehension patterns~\cite{lopes2024language} and replay realistic, time-ordered behavior in interactive simulations~\cite{zhang2024usimagent,park2023generative}, capturing both semantic intent and temporal dynamics. We therefore leverage LLMs as scalable surrogate annotators to infer word-level comprehension properties. This enables automatic construction of the utterance–property pairs that form our action space, overcoming the costly word-level annotation barrier to studying language-based notifications in time-critical settings.

\begin{figure*}
    \centering
    \includegraphics[width=0.75\linewidth]{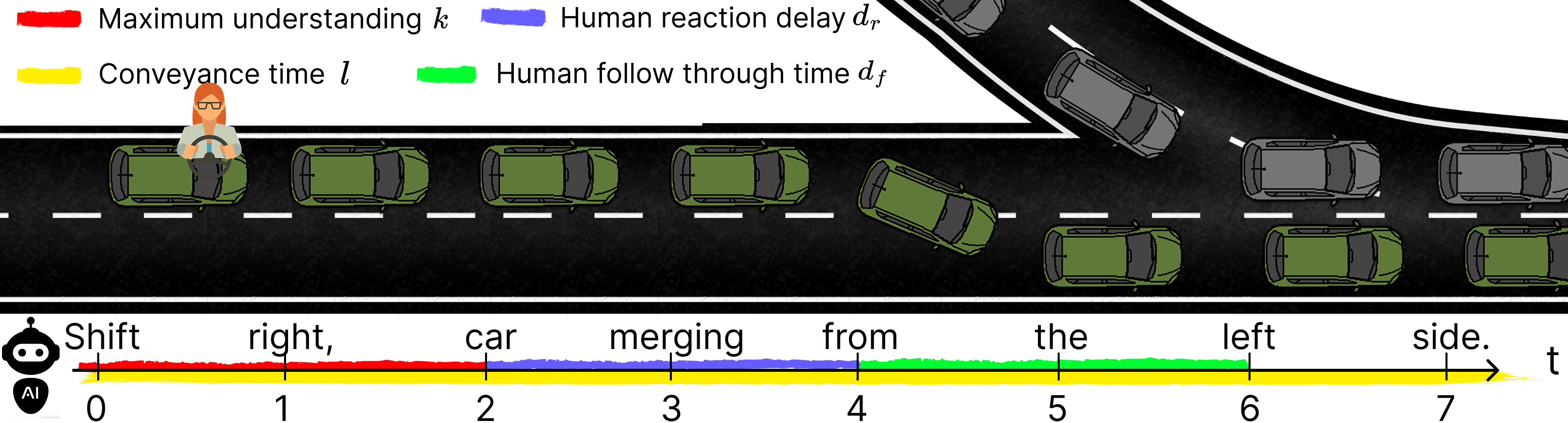}
    \caption{Illustration of a time-critical notification. An AI assistive agent notifies ``Shift right, car merging from the left side'' over $l$ (yellow) time steps. The utterance has communicated actionable information at $k=2$ (red). The human driver then has a reaction delay, $d_r$ (blue), before beginning to move right, lasting the duration of the follow-through time, $d_f$ (green).}
    \label{fig:time-critical}
\end{figure*}

The core contributions of this work are as follows:
\begin{itemize}
    \item \textbf{Formal characterization of language-based notifications in time-critical human–AI collaboration.} We propose a mathematical framework for language-based assistive notifications, formalizing timeliness and informativeness for time-critical settings.
    \item \textbf{Reaction-time–aware decision scheme for time-critical tasks}. We provide a decision-making scheme that explicitly and robustly accounts for critical time delays, while optimizing notification timing and content to maximize safety and task efficiency.
    \item \textbf{Evaluation on various time-critical domains}. We demonstrate the efficacy and robustness of our approach in three domains: piloting, driving, and collaborative cooking.
\end{itemize}

\section{Background and related work}
\label{sec:background}
\subsection{Language-based communications for human-robot collaboration}
Collaboration fluency in language-based human-robot collaboration (HRC) is achieved through balancing effective task-related communication and conversational flow. Some prior works have explored effective communication strategies such as perspective-taking for personalized delivery~\cite{wang2021tom2c}, adapting to human preferences and task states~\cite{buisan2020human}, providing timely reminders~\cite{das2019tarmac}, and correcting false assumptions~\cite{shvo2023proactive, shvo2022resolving}. Other works address timing implicitly through conversational fillers to smooth transitions without delay~\cite{reimann2024survey}. However, these approaches focus on relevance and fluency~\cite{reimann2024survey} rather than explicitly considering the timing of information delivery to support human performance in time-critical tasks.

In time-critical collaboration, the amount of time devoted to communication must be carefully monitored and adapted. While some research acknowledges conveyance duration of assistive household agents~\cite{zhang2023building}, or emphasizes time-related aspects during communication, such as interruption intervals~\cite{unhelkar2020decision}, they still consider communication as an instantaneous or one-timestep~\cite {unhelkar2016contact} action during planning. This simplification neglects a key trade-off: time spent on communication may delay time-critical responses. As a result, there remains a gap in explicitly modeling how communication timing interacts with task execution, especially in domains where even slight delays can lead to failure.

\subsection{Human-robot collaboration in time-critical settings}
We define time-critical HRC as scenarios where the effectiveness of the robot actions depends on the timing of their influence upon the human. Prior work has approached this challenge from several complementary angles. Some studies predict human intent from implicit cues such as gaze, posture, and motion~\citep{breazeal2005effects, hoffman2023inferring}, while others model cognitive delays using drift-diffusion frameworks to capture reaction times in forced-choice tasks~\citep{fudenberg2020testing, mohammad2024driver}, or explore aspects such as engagement~\cite{sundareswara2013using}. At the task level, researchers analyze how robot interventions shape performance and completion time~\citep{chen2022mirror, choudhury2019utility}. In parallel, communication-focused approaches seek to optimize the timing of motion cues~\citep{lee2023effect}, non-verbal signals, and natural language instructions~\citep{baraglia2017efficient} to improve coordination. 
Although timing improves efficiency, these studies typically trade off task-execution effort against communication quantity (e.g., distance cost of motion cues, number of utterances)~\cite{unhelkar2020decision}.

Beyond these trade-offs, other lines of work examine human reaction times in time-critical settings, which focus on modeling reactions to instantaneous cues within well-defined events, often in contexts such as intelligent transportation systems~\cite{mohammad2024driver} or warning systems~\cite{sundareswara2013using, li2024optimal, jin2016optimal, wang2024enhancing}, or designed experiments~\cite{trueblood2011modeling,ratcliff2018modeling}. In contrast, we study language-based communication in time-critical HRC, where the duration of notification delivery can become a critical constraint. Unlike instantaneous warnings, language often conveys richer content to guide human decision-making, creating an underexplored trade-off between timeliness and informativeness.

\subsection{LLMs for modeling human reasoning and behavior}
Modeling human reactions is critical for human–AI collaboration. Large Language Models (LLMs) not only generate language, but also capture aspects of reasoning and behavior, demonstrating abstraction~\cite{zhang2024proagent}, common-sense reasoning~\cite{brown2020language}, decision-making across domains~\cite{klissarov2023motif, mandi2024roco}, and simulating feedback in interactive settings~\cite{park2023generative, aher2023using}. They have further been used as surrogate annotators, often matching or surpassing crowdworkers in structured tasks~\cite{gilardi2023chatgpt, srikanth2025algorithmic}, and have been shown to capture fine-grained predictability and informativeness in comprehension~\cite{lopes2024language}. Together, these results position LLMs as scalable proxies for human reasoning and annotation, motivating use as surrogate human reaction models in our framework.

\section{Problem definition}
\label{sec:problem_definition}
We consider an assistive notification task within a time-critical human-AI collaboration scenario, where the effectiveness of the assistance depends on the timeliness and informativeness of communicated notifications. 
In the scenario, we assume two distinct classes of heterogeneous agents with different objectives~\cite{luebbers2023autonomous}: a human agent that directly affects a physical environment with their actions, and an assistive agent that delivers notifications as verbal utterances to guide or alert the human.
The assistive agent's objective to deliver notifications that not only contain actionable instructions (i.e., content that is sufficiently informative for the human to understand and act upon) but also, when appropriate, those that include additional context on how to carry out the instruction. For instance, to enhance the human's situational understanding or drive them towards a long-term goal.

To effectively provide decision assistance, the assistive agent needs a model of (i) how the human would solve the task unaided and (ii) how the human interprets and responds to notifications. Accordingly, the next subsection introduces a reactive human model that captures baseline task behavior and notification-driven adjustments; we then use this model to train the notifier policy.

\subsection{Reactive human model}
\label{sec:human_model}
Our reactive human model, $H_{react} = (\mathrm{MDP}, \mathcal{M}_{\mathrm{react}})$, contains a decision-making process for completing the task, formulated as a Markov Decision Process (MDP), and an utterance reaction model $\mathcal{M}_{\mathrm{react}}$. 
\subsubsection{Task-completion MDP}
The MDP consists of a state space, $S^h$, action space, $A^h$, a stochastic transition function, $T^h$, describing the model's action-based state transition dynamics, a reward function, $R^h$, and a discount factor $\gamma$. The policy for the task-completion MDP is denoted as $\pi_{MDP}(s^h)$, where $s^h \in S^h$. 
\subsubsection{Utterance reaction model}
We define our utterance reaction model as $\mathcal{M}_{\mathrm{react}} = (U^g, A^h, I, R^h_{react}, \rho^{d_f}_{d_r})$, where $U^g$ is a set of guiding utterances, which each utterance $\mathbf{u}$ is a sequence of $n$ words, i.e., $(u_{t-n}, u_{t-n+1}, ..., u_{t})$, generated by one or more assistive agent notification action $a^g$. The agent's behavior can thus be as $(a^g_{t-n}, \emptyset, ..., a^g_{t})$, where most steps correspond to a null action, and occasional non-null actions produce an utterance spanning multiple words. $I: U^g \rightarrow \mathbb{R}_{\geq 0}$ is a scalar, task-relevant informativeness measure~\cite{10.1162/opmi_a_00188}, that states how much of the notification's actionable content is conveyed to the human. $I(\mathbf{u}) = 0$ indicates no action-independent details (e.g., How/Why that do not change the instructed action), and $I(\mathbf{u}') \geq I(\mathbf{u})$ whenever $\mathbf{u}'$ has more task-relevant details. $R^h_{react}$ is the reaction reward function, and $\rho^{d_f}_{d_r}$ is a reaction function, which updates the intended human action based on the utterances and human reaction delay, $d_r$, and follow-through duration, $d_f$. 

\subsubsection{Task-relevant informativeness}
We use a length-based proxy that measures the action-independent tail of the utterance after it first becomes actionable. Let $l$ be the length of $\mathbf{u}$ and  $I(\mathbf{u}) = l$.
While the effect of informativeness may manifest in different response behavior for different tasks (see Sec.~\ref{sec:informativeness} for details), we model both follow-through duration and reward as a domain-specific function of informativeness, $d_f(I(\mathbf{u}))$ and $R^h_{react}(I(\mathbf{u}))$, respectively. For example, in the driving domain, $d_f$ is a monotonic increasing function of $l$; for the piloting domain, we performed both constant (Appendix~\ref{app:hard_ll_informativenss}) and linear (Appendix~\ref{app:notifier_polices}) mappings. A brief message like ``Slow down!'' could quickly cause the human to slow, but with indefinite response, while a longer instruction, such as ``Slow down for the next few seconds'' reveals more information, such as how to perform the instruction, which leads to a more targeted response, but only after some delay. Refer to Sec.~\ref{sec:exp:q3} and Appendix~\ref{app:hard_ll_informativenss} for details on $R^h_{react}(I(\mathbf{u}))$.
(See Appendix~\ref {app:informative_content_types} for discussion on different types of informative notifications.)

\begin{figure*}[ht]
    \centering
    \includegraphics[width=0.85\linewidth]{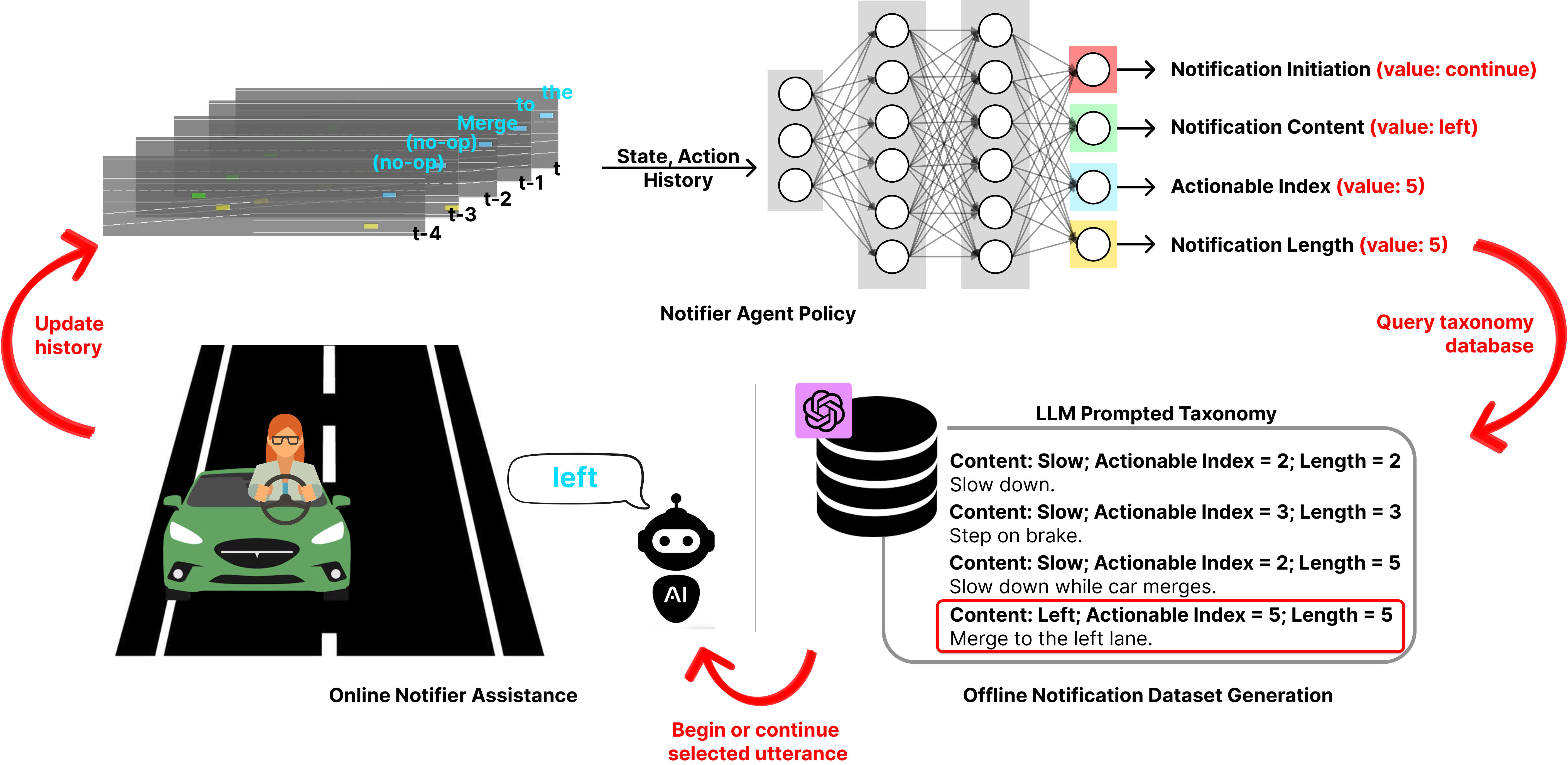}
    \caption{We present a notifier assistant that learns to provide timely information to help a human solve a task. The notifier receives recent state-action history to predict whether to provide a notification and properties the desired notification (content, desired reaction time, and length), which is then used to retrieve a precomputed utterance from the offline-generated utterance database matching these criteria and delivers it to the human.}
    \label{fig:overview}
\end{figure*}

\subsubsection{Types of notifications}
Since we focus on notifications that are actionable and informative, we associate 
$a^g$ with a property tuple $(c, k, l)$, where $c$ represents the content topic, $k$ is the \textit{comprehension time}~\cite{devault2009can, devault2011incremental}, i.e. the time relative to the beginning of the utterance at which the human initiates their reaction as actionable information is conveyed, where $k \le l$, and $l$ denotes the \textit{conveyance time}, i.e. the time that the notification takes to complete. Note that all times in this paper are expressed in terms of words, where the expected duration of each word is approximately 0.3 seconds~\cite{yuan2006towards}.

\subsubsection{Timing latencies}
In addition to comprehension time, our human reaction model includes two temporal latencies: the reaction delay $d_r$, the interval between comprehension (at time $k$) and initiation of the instructed action, parameterized using empirical reaction-time estimates from prior work \cite{drozdziel2020drivers}, and the follow-through duration $d_f$, the period for which the action is sustained once understood \cite{gollwitzer1999implementation} (see Fig.~\ref{fig:time-critical}).

\subsubsection{Reaction function}
The reaction function $\rho^{d_f}_{d_r}$ outputs the human reaction behavior after the notification is comprehended at time step $t+k$. Here, $t$ is the absolute start of the notification action $a^g = (c,k,l)$, $k$ is the comprehension delay relative to $t$, followed by a reaction delay $d_r$, and sustained for a duration of $d_f$ (see Fig.~\ref{fig:time-critical} for an example). Let $\Delta t$ denote the elapsed time since the notification was comprehended, $t'-(t+k)$, where $t'$ is the absolute current time step. The human reaction function is defined as follows:
\begin{equation}
    \begin{split}
        &\rho^{d_f}_{d_r} (\pi_{MDP}(s_{t'}), c , \Delta t)\\
        & = \begin{cases}
            f(c) &  d_r \le \Delta t \le d_r + d_f \\[3pt] \pi_{MDP}(s_{t'}) & otherwise,
        \end{cases}
    \end{split}
\end{equation}
where $s_{t'} \in S^h$, and $f(c)$ maps the topic to the corresponding reaction $a^h$.

\subsection{Assistive notifier model}
We consider an assistive notification problem in a time-critical human-AI collaboration scenario, formulated as a sequential decision-making task. 
The environment consists of a model of the physical world and of the human.  As physics and human states can evolve \textit{while} an utterance is produced, it may be cast as an augmented-state Markov Decision Process (MDP)~\cite{sutton1998reinforcement}.
Following the convention from prior literature~\cite{chen2020delay, chen2021delay}, we define the augmented state space as a finite history of windowed state-action pairs. 
Hence, we turn the delayed-feedback task into a standard MDP, a tuple $(\mathcal{X}, \mathcal{A}, T, R, \gamma)$, with the augmented state space $\mathcal{X} = (S \times \mathcal{A})^n$, $S= S^w \times S^h \times S^g$, where $S^w$, $S^h$, and $S^g$ are the environment, human, and assistive notifier agent state spaces respectively, $\mathcal{A}$ is the action space of the assistive notifier, and $n$ is the history window length. 

At each timestep, the notifier takes an action $a^g_t \in \mathcal{A}$, which is either null (no notification) or a notification characterized by $(c,k,l)$. If a notification is issued, the reaction function $\rho^{d_f}_{d_r}$, defined in Sec.~\ref{sec:human_model}, determines $a^h$ by combining the utterance properties with temporal dynamics (i.e., $d_r$, $d_f$, and $\Delta t$).
The state transition then decomposes as 
\begin{equation}
    T(x_{t+1}|x_t,a^g_t) = \sum_{a^h \in A^h} P(a^h|x_t,a^g_t) P_{\mathrm{env}}(x_{t+1}|x_t,a^h),
\end{equation}
where $P(a^h|x_t,a^g_t)$ is induced by $\rho^{d_f}_{d_r}$, and $P_{\mathrm{env}}$ captures the environment given the realized human action. The immediate reward $R(x_{t+1},a^g_t)$ balances the task improvement and communication costs (e.g., penalties for uttering notifications). The objective is to learn an optimal notification policy $\pi^*: \mathcal{X}\rightarrow \mathcal{A}$ that maximizes the expected discounted cumulative reward.

\section{Language-based reaction-aware notifier}
\label{sec:reaction_aware_notifier}
To address the timeliness and informativeness in assistive notification, we propose a \textit{Language-Based Reaction-Aware Notifier} that learns to jointly reason over both \textit{when} to communicate and \textit{what} to communicate. 
The notifier comprises two interdependent components: (1) a timing-and-intent network that determines the appropriate time to deliver a notification and selects a semantic intent or category for the message, accounting for context, human comprehension time, and reaction delays; and (2) a message realization module, which queries an offline taxonomy database to retrieve utterance that best matches the selected semantic intent. This approach enables adaptive, context-aware notification delivery that is both behaviorally informed and capable of operating in real time.

\subsection{Timing-and-intent network}
Our timing-and-intent network is designed to account for both human reaction delays and the temporal dynamics of message conveyance. To address these challenges, we draw on delay-aware reinforcement learning literature~\citep{chen2020delay, chen2021delay}, where the agent's observation is augmented with the most recent $n$ state–action pairs. This history stack~\citep{mnih2013playing} formulation provides the temporal context needed to reason about decision-to-effect delays and to preempt mid-utterance by continuing, truncating, or replacing a message as conditions evolve.

For the policy architecture, prior work shows that in rapid-decision regimes heavier encoders add inference cost with little benefit and that simple frame-stacked inputs yield stable learning in discrete, delay-aware control \citep{anokhin2025handling}. Guided by this, we use a lightweight MLP over the history stack and train with Proximal Policy Optimization \citep{schulman2017proximal}. See Appendix~\ref{app:notifier_architecture} for further architecture details. 

\subsection{Offline taxonomy generation via an LLM-surrogate reaction model}
Given the optimal notification action $a^g_t$ from the timing-and-intent network, we aim to provide utterances that correspond to the property of $a^g_t$. Such a process involves modeling human reactions at the word level. To construct the taxonomy database in a scalable way, 
we use LLMs as surrogate models of human reactions to notification utterances. This allows us to assign word-by-word labels to notifications for a given domain, grounded in expected human response. 

\begin{tcolorbox}[
  colback=blue!5!white,
  colframe=blue!40!black,
  title= Sample Prompt: Rating Comprehension Progression,
  fonttitle=\small,
  boxrule=0.3mm,
  left=2mm,
  right=2mm,
  top=1mm,
  bottom=1mm,
  boxsep=0.5mm,
  enhanced,
  fontupper=\small,
]

Given a pilot instruction, estimate a human's comprehension level of the intended action word-by-word by \\
(1) Starting at 0\% comprehension \\
(2) Reading each word left to right, updating comprehension: \\
- Key action words (e.g., “Slow”) boost comprehension\\
...\\
Output a list of comprehension values (0–100\%) after each word per instruction.\
\tcbline

Example (``Slow down'' type):\\
Immediate $|$ speed $|$ reduction $|$ needed $|$ danger! \\
5\% $|$ 20\% $|$ 80\% $|$ 80\% $|$ 100\%\\
Adjust $|$ speed $|$ prepare $|$ to $|$ avoid $|$ the $|$ zone. \\
5\% $|$ 50\% $|$ 50\% $|$ 50\% $|$ 70\% $|$ 70\% $|$ 70\%
\end{tcolorbox}

We build on LLMs' ability to approximate human comprehension~\citep{srikanth2025algorithmic, chatgptannotator} and prompt LLMs to: (1) generate notifications for a topic $c$, and (2) assign comprehension progression scores to each word~\citep{zheng2023judging}, identifying the word index $k$ where a human might react~\cite{devault2011incremental}. 
The generated notifications are then categorized by their properties. A sample prompt for (2) is shown above; see Appendix~\ref{app:noti_analysis} for more details.  For this paper, these numeric comprehension sequences are illustrative, and we leave calibration to actual humans as future work.

\begin{table*}[ht!]
\centering
\resizebox{0.8\textwidth}{!}{%
\begin{tabular}{l|cc||cccc}
\toprule
\multirow{2}{*}{\textbf{Notifier Policy}} 
& \multicolumn{2}{c||}{\textbf{Success Rate (↑)}} 
& \multicolumn{4}{c}{\textbf{Secondary Metrics}} \\
\cmidrule(lr){2-3}\cmidrule(lr){4-7}
& \textbf{Piloting} & \textbf{Driving} 
& \textbf{Piloting Noti.} & \textbf{Piloting Follow} 
& \textbf{Driving Noti.} & \textbf{Driving Follow} \\
\midrule
Heuristic
  & 0.00               & N/A               
  & 0.50               & 0.76               
  & N/A                & N/A                \\

Delay-Free Notifier
  & $0.22 \pm 0.04$    & $0.28 \pm 0.03$    
  & $0.25 \pm 0.01$    & $0.83 \pm 0.03$ 
  & $0.58 \pm 0.18$    & $0.04 \pm 0.02$    \\

Notifier w/ Convey
  & $0.94 \pm 0.03$    & $0.87 \pm 0.11$    
  & $0.22 \pm 0.01$ & $0.82 \pm 0.01$    
  & $0.22 \pm 0.10$ & $0.32 \pm 0.13$ \\

Convey \& React (Ours)
  & \cellcolor{best}$0.97 \pm 0.02$ & \cellcolor{best}$0.93 \pm 0.02$ 
  & $0.22 \pm 0.01$ & $0.82 \pm 0.01$    
  & $0.35 \pm 0.19$    & $0.18 \pm 0.14$    \\
\bottomrule
\end{tabular}%
}
\caption{Success rates for piloting and driving (primary metric) with secondary metrics, notification frequency and follow-through rate, illustrating trade-offs from considering human reaction delays.  Blue indicates the best success rate.}
\label{tab:exp:piloting_driving}
\end{table*}

\begin{table*}[ht]
\centering
\footnotesize
\resizebox{0.9\textwidth}{!}{%
\begin{tabular}{l|
>{\columncolor{white}}ccc|
>{\columncolor{white}}ccc|
>{\columncolor{white}}ccc}
\toprule
\multirow{2}{*}{\textbf{Reaction}} 
& \multicolumn{3}{c|}{\textbf{Avg.\ Noti.\ Freq.\ ($\downarrow$)}} 
& \multicolumn{3}{c|}{\textbf{Avg.\ Follow-Through Rate ($\uparrow$)}} 
& \multicolumn{3}{c}{\textbf{Success Rate (\%) ($\uparrow$)}}\\
\cmidrule(r){2-4}\cmidrule(r){5-7}\cmidrule{8-10}
& \(\mathcal{N}(2,0.5)\) & \(\mathcal{N}(2,1.0)\) & Matching 
& \(\mathcal{N}(2,0.5)\) & \(\mathcal{N}(2,1.0)\) & Matching 
& \(\mathcal{N}(2,0.5)\) & \(\mathcal{N}(2,1.0)\) & Matching \\
\midrule
\(d_r = 0\)  
  & \cellcolor{best}$0.21\pm0.00$ & \cellcolor{best}$0.21\pm0.01$ & $0.22\pm0.01$ 
  & $0.68\pm0.03$ & $0.69\pm0.06$ & $0.82\pm0.01$ 
  & \cellcolor{best}$0.97\pm0.01$ & \cellcolor{best}$0.96\pm0.02$ & $0.94\pm0.03$ \\

\(d_r = 1\)  
  & \cellcolor{best}$0.21\pm0.01$ & \cellcolor{best}$0.21\pm0.01$ & $0.22\pm0.01$ 
  & $0.68\pm0.04$ & $0.68\pm0.06$ & $0.71\pm0.05$ 
  & \cellcolor{best}$0.98\pm0.00$ & \cellcolor{best}$0.98\pm0.01$ & $0.98\pm0.01$ \\

\(d_r = 2\)  
  & \cellcolor{best}$0.21\pm0.01$ & \cellcolor{best}$0.21\pm0.01$ & $0.22\pm0.01$ 
  & \cellcolor{best}$0.81\pm0.01$ & \cellcolor{best}$0.83\pm0.01$ & $0.82\pm0.01$ 
  & \cellcolor{best}$0.96\pm0.02$ & \cellcolor{best}$0.96\pm0.02$ & $0.97\pm0.02$ \\

\(d_r = 3\)  
  & $0.42\pm0.02$ & $0.42\pm0.00$ & $0.27\pm0.01$ 
  & $0.04\pm0.02$ & $0.05\pm0.05$ & $0.88\pm0.01$ 
  & $0.00\pm0.00$ & $0.00\pm0.00$ & $0.11\pm0.06$ \\

\(d_r = 4\)  
  & $0.46\pm0.00$ & $0.46\pm0.00$ & $0.15\pm0.01$ 
  & $0.01\pm0.00$ & $0.01\pm0.02$ & $0.28\pm0.01$ 
  & $0.00\pm0.00$ & $0.00\pm0.00$ & $0.10\pm0.02$ \\
\bottomrule
\end{tabular}%
}
\caption{Notifier policy robustness to human reaction delays $d_r$ (timesteps) in the Lunar Lander domain. The policy remains robust for $d_r$ values below the training value. Blue highlights indicate performance that either matches or outperforms the Matching condition (human $d_r$ aligned with training $d_r$).}
\label{tab:robustness}
\end{table*}

\section{Experiments}
\label{sec:experiments}
\subsection{Domains}
\label{sec:domains}
We evaluate our approach in time-critical domains that lend themselves to assistance, where the human lacks certain task-relevant information that the assistant can access. We modify three domains for this purpose: Lunar Lander~\cite{brockman2016openai}, Highway Merging~\cite{highway-env}, and Steakhouse~\cite{hsu2025integrating}.  In each, the human is under time pressure, and lacks certain information, offering complementary testbeds for evaluating notification design.

\subsubsection{Lunar Lander (Piloting)} This domain~\cite{brockman2016openai} involves a human piloting a spacecraft to land on a designated platform, while navigating a dynamic environment containing danger zones, which are visible to the notifier but occluded from the pilot. Due to the rapidly evolving and control-intensive nature of the task, the notifier must provide timely notifications, helping the pilot quickly adjust to stay outside of danger zones, remain in control of the spacecraft, and land safely.

\subsubsection{Highway Merging (Driving)} This domain, modified from \citep{highway-env}, features a human-driven ego vehicle on a highway with three merging events, adapted from \citep{highway-env}. Merging vehicles are unaware of the ego vehicle, while the human driver expects them to yield. The notifier's objective is to proactively alert the driver to maneuver away from impending collision risks, thereby encouraging smooth traffic flow free of unnecessary decelerations or abrupt maneuvers. 

\subsubsection{Steakhouse (Cooking)} This domain~\cite{hsu2025integrating}, adapted from~\cite{carroll2019utility}, simulates a community kitchen where a human collaborates with a notifier to complete subtasks (e.g., cooking meat, chopping radishes, washing dirty plates, and plating cooked steak garnished with chopped radishes). The human has partial observability—some stations are already occupied and only revealed upon approach—while the notifier has full observability. The notifier proactively issues concise prompts, ``On your left'', to encourage quick action interventions, or incremental actionable notifications like ``Shift left, all the stations are occupied'', to enable quick interventions and support the human’s mental model. This setup illustrates how timely context-aware notifications aid both immediate actions and broader decision-making.

\subsection{Experimental setup}
\label{sec:experiment_setup}

\subsubsection{Notifier Architecture}
The notifier uses a feedforward neural network trained on domain-specific processed observations (details in Appendix~\ref{app:notifier_architecture}) and outputs separate categorical distributions for (1) notification initiation, (2) topic $c$, (3) comprehension time $k$, and (4) length $l$, each predicted through a distinct linear projection from shared features. The initiation projection determines whether to notify, while the remaining projections define the notification action $a^g = (c, k, l)$.

\subsubsection{Notification types} 
\label{sec:noti_types}
Apart from setting the notification action as above, we consider three notification action types, each defining different bounds of the action space $\mathcal{A}$. \textit{Complete-utterance notifications} assume the human comprehends only after the entire notification is delivered ($k = l$) (e.g., ``Once past the merge point slow down.''). \textit{Topic-only notifications} convey minimal, high-urgency commands without elaboration (e.g., ``Stop!''). \textit{Incrementally actionable notifications} are designed for partial comprehension during delivery ($k \leq l$) that is sufficient to trigger a response, with the remaining content adding clarifying context. For instance, ``Slow down to avoid the left merging vehicle'' prompts action upon hearing ``Slow down,'' with the rest of the notification providing context.

\subsubsection{Notification informativeness}
\label{sec:informativeness}
Across all domains, the informativeness of a notification $\mathbf{u}$ is assumed to be a monotone non-decreasing function of its message length $l$, i.e., $I(\mathbf{u}) = f(l)$ with $f$ monotone. Its effects manifest differently depending on the task: in piloting, informativeness improves task understanding and stability, which we capture by defining the human reward as $R^h_{react}(I(\mathbf{u}))$; in driving, it modulates reaction behavior through follow-through duration $d_f(I(\mathbf{u}))$, where more informative notifications sustain actions for longer; and in cooking, it conveys station occupancy information, thereby updating the human’s environment state and shaping subsequent decision-making.

\subsubsection{Human agent} 
The task-completion MDP policy is trained separately for each domain to capture environment-specific dynamics (Appendix~\ref{app:human_agent_polices}). In contrast, all notifiers are evaluated under identical parameter settings for the reaction function $\rho_{d_r}^{d_f}$, ensuring consistent notification–reaction dynamics. The only exception is the robustness analysis study (Section~\ref{sec:exp_robustness}), where reaction parameters are varied to examine model mismatch.

\subsubsection{Evaluation metrics}
Our primary evaluation metric is success rate, which captures both task completion and safety across domains. To further characterize notification behavior, we report three secondary metrics: \textit{notification frequency} (average notifications per episode), \textit{follow-through rate} (the proportion of delivered notifications that the human comprehends and acts upon), and \textit{long notification rate} (the proportion of notifications that include extended, context-rich content). Together, these metrics quantify both the effectiveness and efficiency of different notification strategies.

\subsubsection{Baselines}
We compare our method against three baselines that differ in how they account for notification timing, while holding the human reaction model fixed across domains. This isolates the impact of explicitly modeling conveyance and reaction delays in the notifier’s decision-making.
\begin{itemize}
    \item \textbf{Heuristic Notifier}: Issues alerts when the human approaches a danger zone, directing them to follow a predefined safe path once a distance threshold is crossed.
    \item \textbf{Delay-Free Notifier (RL) \cite{liu2024llm}}: Assumes notifications are delivered and acted upon instantaneously ($k=0$, $d_r=0$), ignoring both conveyance and reaction delays.
    \item \textbf{Notifier With Conveyance Time Awareness (RL) \cite{zhang2023building}}: Models the time required to deliver a notification ($k>0$), but assumes immediate human response ($d_r=0$).
\end{itemize}

\subsubsection{Our method} \textbf{Notifier With Conveyance and Reaction Time Awareness}: Incorporates both notification conveyance duration and human reaction delays, explicitly modeling a cumulative two-timestep delay ($d_r=2$)~\cite{drozdziel2020drivers}.

\subsection{Results}
We organize our experiments to answer the following:
\begin{itemize}[align=left]
    \item[(Q1)] Does a conveyance and reaction time aware notifier result in significant performance improvements over baseline approaches? 
    \item[(Q2)] Can a notifier trained on a limited distribution of reaction times generalize to out-of-distribution cases?
    \item[(Q3)] Does conveyance and reaction time aware notifier balance notification timeliness and informativeness to improve overall task performance?
\end{itemize}

\begin{table*}[ht]
\centering
\footnotesize
\resizebox{\textwidth}{!}{%
\begin{tabular}{l|cc||cc|cc|cc}
\toprule
\multirow{2}{*}{\textbf{Notifier Policy}}
  & \multicolumn{2}{c||}{\textbf{Success Rate (\%) ↑}}
  & \multicolumn{2}{c|}{\textbf{Avg.\ Noti.\ Freq.}}
  & \multicolumn{2}{c|}{\textbf{Avg.\ Follow-Through Rate}}
  & \multicolumn{2}{c}{\textbf{Long Noti. Rate}} \\
\cmidrule(r){2-3}\cmidrule(lr){4-5}\cmidrule(l){6-9}
  & \textbf{Piloting} & \textbf{Driving}
  & \textbf{Piloting} & \textbf{Driving}
  & \textbf{Piloting} & \textbf{Driving}
  & \textbf{Piloting} & \textbf{Driving} \\
\midrule
Topic Only (\(\langle c\rangle\))
  & \cellcolor{best}$0.97\pm0.02$  & $0.93\pm0.04$
  & \(0.22\pm0.01\)          & \(0.20\pm0.09\)
  & \(0.77\pm0.02\)          & \(0.32\pm0.09\)
  & \(N/A\)          & \(N/A\)\\

Complete-Utterance (\(\langle c,l\rangle\))
  & \cellcolor{best}$0.97\pm0.02$ & \cellcolor{best}$0.97\pm0.03$
  & \(0.22\pm0.01\)          & \(0.37\pm0.12\)
  & \(0.82\pm0.01\)     & \(0.29\pm0.04\) 
  & \(0.01 \pm 0.01\) &  \(0.37 \pm 0.23\)\\
\bottomrule
\end{tabular}%
}
\caption{Performance of different notification types in piloting and driving. Success rates are shown first as the primary metric; notification frequency and follow‐through rate are secondary metrics.  Blue indicates the best success rate.}
\label{tab:results-notification-type}
\end{table*}

\subsubsection{Performance analysis (Q1)}

We evaluate notifier effectiveness primarily using success rate, which reflects both task safety and goal achievement in time-critical settings. As shown in Table~\ref{tab:exp:piloting_driving}, our proposed Convey \& React notifier—explicitly modeling both conveyance and reaction delays—achieves the highest success rates in piloting (0.97) and driving (0.93). By contrast, the Delay-Free baseline performs poorly, highlighting the risks of ignoring timing altogether. In between, the w/ Convey baseline, which accounts only for conveyance delays, already achieves strong performance, demonstrating that modeling this factor alone yields substantial gains. Building on this, incorporating reaction delays provides further consistent improvements across domains. Overall, these results show that explicitly modeling both conveyance and reaction delays is essential for maximizing performance in time-critical assistive settings.

\begin{figure}[ht]
    \centering
    \includegraphics[width=\linewidth]{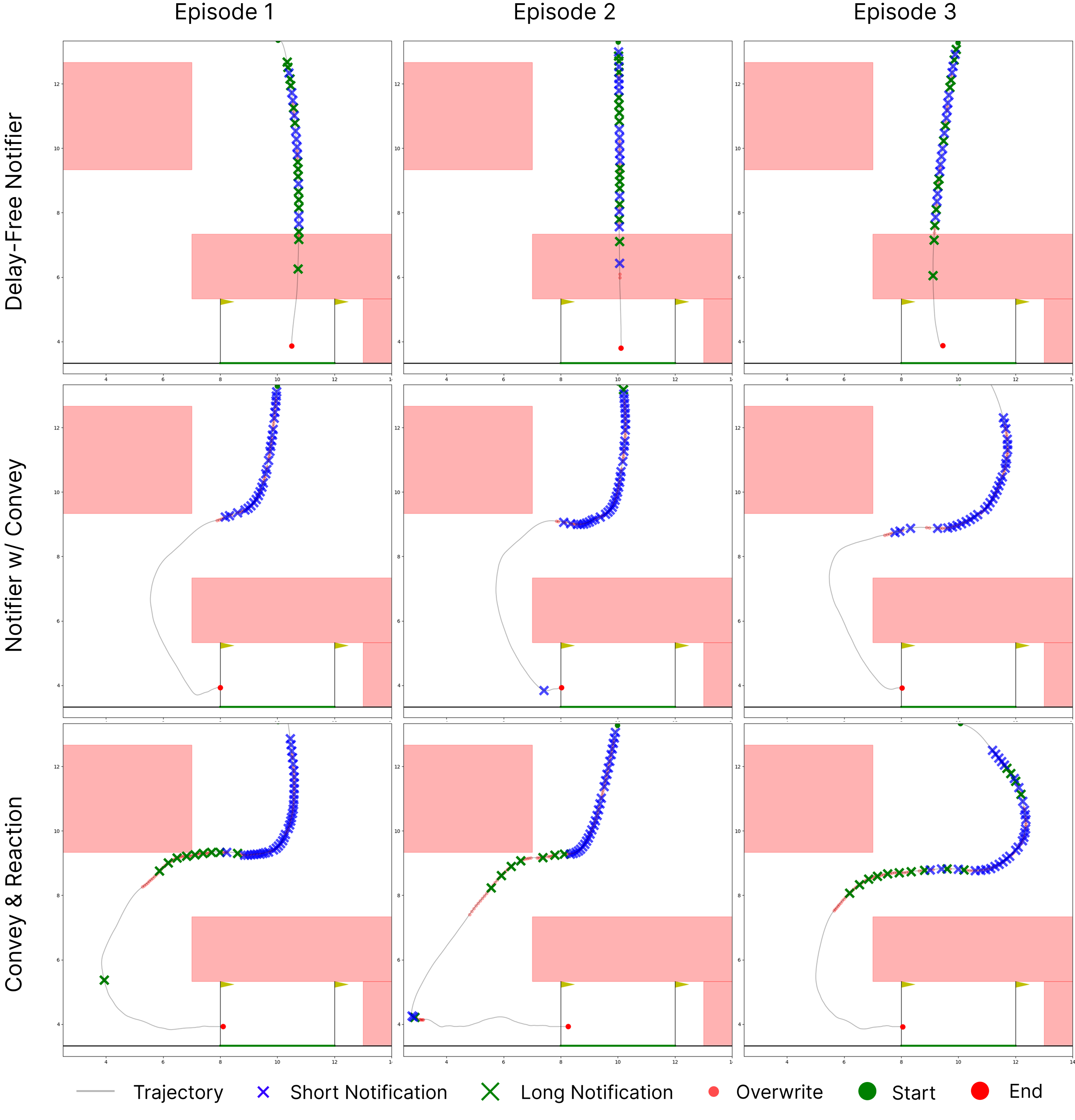}
    \caption{Trade-off between notification timeliness and informativeness across policies in the Lunar Lander (Harder Version) domain. Longer notifications provide higher rewards but take more time to convey, which may be too late in fast-changing environments. The Delay-Free Notifier exemplifies this failure. The Notifier w/ Convey issues short alerts to avoid danger zones, while Convey \& React begins with short notifications and switches to longer ones once the lander stabilizes above them. Both stop notifying once the lander is safe to descend unaided.}
    \label{app:fig:hard_ll_traj}
\end{figure}

\begin{figure*}[ht]
    \centering
    \includegraphics[width=0.95\linewidth]{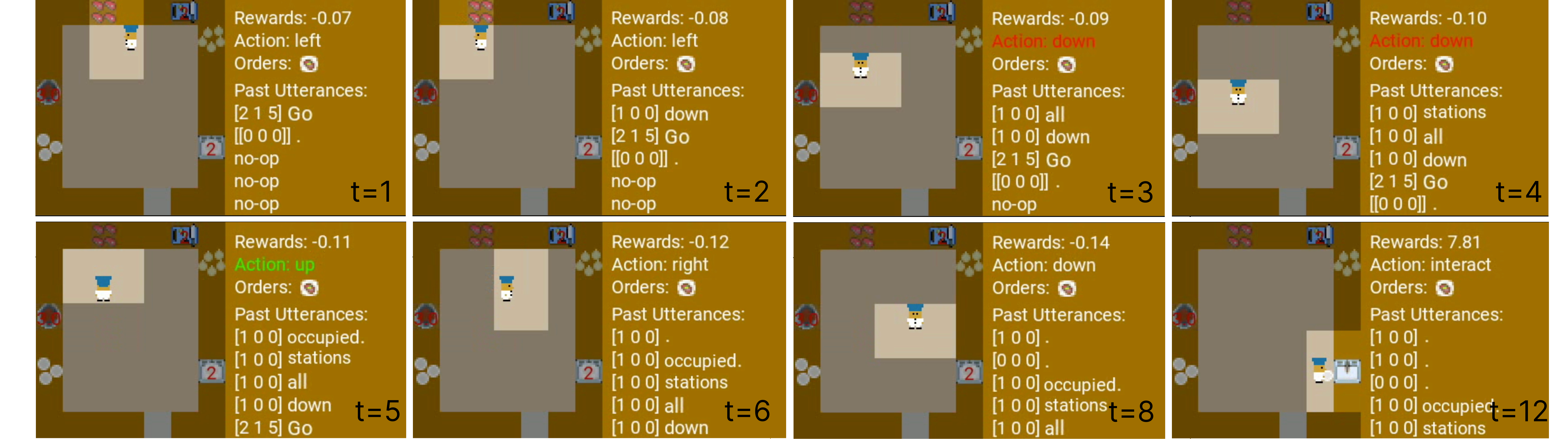}
    \caption{In Steakhouse domain, the notification ``Go down, all stations occupied'' (denoted [2 1 5] in the figures) begins at t = 0. Mid-sentence, the human comprehends the initial instruction and starts moving downward. By t = 5, upon full delivery and comprehension, the human updates their mental model and proceeds with optimal actions based on the current kitchen state.}
    \label{fig:steakhouse}
\end{figure*}

\subsubsection{Robustness analysis (Q2)}
\label{sec:exp_robustness}
Human responses to notifications vary due to differences in comprehension, reaction delay, and follow-through. In our formulation, these sources of variability enter through the reaction function $\rho^{d_f}_{d_r}$, which determines the lag between a notification and observed behavior. For tractability, we vary the reaction delay $d_r$, since robustness to unseen delays directly tests generalization to out-of-distribution timing, while variability in comprehension or follow-through would enter through the same functions ($\rho^{d_f}_{d_r}$ or $d_f(I(\mathbf{u}))$) and are left for future work.

Table~\ref{tab:robustness} compares two training regimes: a \textit{population-trained} notifier, which samples $d_r$ from a Gaussian distribution, and a \textit{matching} notifier trained with a fixed delay equal to evaluation (upper bound). The population-trained notifier generalizes well to unseen delays shorter than those sampled during training, achieving success rates near the upper bound (0.96–0.98). However, both population-trained and matching policies' performance drop once delays become very long ($d_r \geq 3$), reflecting a fundamental limitation: excessively delayed human reactions cannot be fully compensated by notification policies.

\subsubsection{Timeliness and informativeness analysis (Q3)}
\label{sec:exp:q3}
We examine how notifiers balance timeliness and informativeness by comparing Topic-Only notifications—short prompts that trigger immediate actions (e.g., “Speed up!”)—and Complete-Utterance notifications—longer messages that sustain actions (e.g., “Speed up for 5 seconds”). In our first experiment, informativeness is linked to follow-through duration $d_f(I(u))$ (Sec.~\ref{sec:human_model}). Table~\ref{tab:results-notification-type} shows that in the piloting domain, both notification types reach a success rate of 0.97, showing that highly dynamic environments leave little room for sustained actions. In the driving domain, however, Complete-Utterance improves success (0.97 vs. 0.93) by sustaining driver responses. Although success rate shows a benefit, secondary metrics reveal a trade-off: longer utterances are often interrupted by state changes, which forces preemption and lowers follow-through on notifications that were not fully delivered.

The saturated success rate in piloting suggests that informativeness, when defined through follow-through, has a limited effect in highly dynamic environments, with benefits more likely reflected in dimensions not captured by success rate, such as communication efficiency, collaboration fluency, or trust. Hence, we run a second experiment where informativeness is tied to reward rather than follow-through, allowing us to observe how policies reallocate toward longer messages while holding compliance fixed. Additionally, we study a harder version of the original lunar lander environment (see Appendix~\ref{app:ll_hard} for details) to further evaluate the trade-off. Figure~\ref{app:fig:hard_ll_traj} shows that the Delay-Free notifier issues long utterances too late, while the w/ Convey notifier improves safety with short, timely alerts but, because assuming zero reaction delay, avoids longer messages that risk late interventions. In contrast, our Convey \& React notifier adapts by beginning with short alerts to prevent immediate failure, then switching to longer utterances once the lander stabilizes, and stopping once passed danger zones.

Together, these results demonstrate that explicitly modeling both conveyance and reaction delays enables notifiers to adapt notification length to task conditions—favoring timeliness when urgent and informativeness when time permits.

\subsubsection{Incrementally actionable notifications} 
\label{sec:incremental_notifications}
We also demonstrate incrementally actionable notifications in the Steakhouse cooking domain (Fig.~\ref{fig:steakhouse}), where notifications are structured with an initial action cue followed by additional context. For example, the utterance “Go down, all stations occupied” begins at $t=0$, by $t=2$ the human has already acted on the initial cue (“Go down”), and by $t=5$ the full message (“all stations occupied”) updates the human’s situational awareness for future decisions. This demonstration shows how incrementally actionable notifications can prevent immediate errors while providing context that supports longer-term planning. Although they offer limited benefit in highly dynamic domains such as piloting or driving, they highlight the potential value of informativeness in domains like assistive cooking, where success depends on balancing short-term responses with future coordination.

\section{Conclusion and Discussion}
In this paper, we proposed a framework for optimizing the timing and content of language-based notifications in AI-assisted interactions, explicitly modeling instruction duration, human comprehension delay, and subsequent response. By formulating the problem as a sequential decision-making task and applying reinforcement learning, our method outperformed baselines that assume instantaneous human reactions or account only for notification delivery time. Although our evaluation currently relies on LLMs to simulate comprehension and draws from prior work on reaction time delays, validating these assumptions with human subjects and developing adaptive mechanisms to handle varying reaction times (e.g., due to fatigue) remain important directions for the community. Another promising avenue is refining the offline taxonomy dataset with human-in-the-loop feedback. Overall, our results show that accounting for both informativeness and timeliness is not only feasible but impactful, and this work is expected to serve as a foundation for more human-aligned, adaptive communication in time-critical settings.

\bibliography{references}  

\begin{thebibliography}{60}
\providecommand{\natexlab}[1]{#1}

\bibitem[{Aher, Arriaga, and Kalai(2023)}]{aher2023using}
Aher, G.~V.; Arriaga, R.~I.; and Kalai, A.~T. 2023.
\newblock Using large language models to simulate multiple humans and replicate human subject studies.
\newblock In \emph{International Conference on Machine Learning}, 337--371. PMLR.

\bibitem[{Anokhin et~al.(2025)Anokhin, Rishav, Riemer, Chung, Rish, and Kahou}]{anokhin2025handling}
Anokhin, I.; Rishav, R.; Riemer, M.; Chung, S.; Rish, I.; and Kahou, S.~E. 2025.
\newblock Handling Delay in Real-Time Reinforcement Learning.
\newblock \emph{arXiv preprint arXiv:2503.23478}.

\bibitem[{Baraglia et~al.(2017)Baraglia, Cakmak, Nagai, Rao, and Asada}]{baraglia2017efficient}
Baraglia, J.; Cakmak, M.; Nagai, Y.; Rao, R.~P.; and Asada, M. 2017.
\newblock Efficient human-robot collaboration: when should a robot take initiative?
\newblock \emph{The International Journal of Robotics Research}, 36(5-7): 563--579.

\bibitem[{Breazeal et~al.(2005)Breazeal, Kidd, Thomaz, Hoffman, and Berlin}]{breazeal2005effects}
Breazeal, C.; Kidd, C.~D.; Thomaz, A.~L.; Hoffman, G.; and Berlin, M. 2005.
\newblock Effects of nonverbal communication on efficiency and robustness in human-robot teamwork.
\newblock In \emph{International conference on intelligent robots and systems}, 708--713. IEEE.

\bibitem[{Brockman et~al.(2016)Brockman, Cheung, Pettersson et~al.}]{brockman2016openai}
Brockman, G.; Cheung, V.; Pettersson, L.; et~al. 2016.
\newblock {OpenAI} Gym.
\newblock \emph{arXiv preprint arXiv:1606.01540}.

\bibitem[{Brown et~al.(2020)Brown, Mann, Ryder, Subbiah, Kaplan, Dhariwal, Neelakantan, Shyam, Sastry, Askell et~al.}]{brown2020language}
Brown, T.; Mann, B.; Ryder, N.; Subbiah, M.; Kaplan, J.~D.; Dhariwal, P.; Neelakantan, A.; Shyam, P.; Sastry, G.; Askell, A.; et~al. 2020.
\newblock Language models are few-shot learners.
\newblock \emph{Advances in Neural Information Processing Systems}, 33: 1877--1901.

\bibitem[{Buisan, Sarthou, and Alami(2020)}]{buisan2020human}
Buisan, G.; Sarthou, G.; and Alami, R. 2020.
\newblock Human aware task planning using verbal communication feasibility and costs.
\newblock In \emph{International Conference on Social Robotics}, 554--565. Springer.

\bibitem[{Carroll et~al.(2019)Carroll, Shah, Ho, Griffiths, Seshia, Abbeel, and Dragan}]{carroll2019utility}
Carroll, M.; Shah, R.; Ho, M.~K.; Griffiths, T.; Seshia, S.; Abbeel, P.; and Dragan, A. 2019.
\newblock On the utility of learning about humans for human-ai coordination.
\newblock \emph{Advances in Neural Information Processing Systems}, 32.

\bibitem[{Chen et~al.(2021)Chen, Xu, Li, and Zhao}]{chen2021delay}
Chen, B.; Xu, M.; Li, L.; and Zhao, D. 2021.
\newblock Delay-aware model-based reinforcement learning for continuous control.
\newblock \emph{Neurocomputing}, 450: 119--128.

\bibitem[{Chen et~al.(2020)Chen, Xu, Liu, Li, and Zhao}]{chen2020delay}
Chen, B.; Xu, M.; Liu, Z.; Li, L.; and Zhao, D. 2020.
\newblock Delay-aware multi-agent reinforcement learning for cooperative and competitive environments.
\newblock \emph{arXiv preprint arXiv:2005.05441}.

\bibitem[{Chen, Fong, and Soh(2022)}]{chen2022mirror}
Chen, K.; Fong, J.; and Soh, H. 2022.
\newblock MIRROR: Differentiable Deep Social Projection for Assistive Human-Robot Communication.
\newblock In \emph{Proceedings of Robotics: Science and Systems}.

\bibitem[{Choudhury et~al.(2019)Choudhury, Swamy, Hadfield-Menell, and Dragan}]{choudhury2019utility}
Choudhury, R.; Swamy, G.; Hadfield-Menell, D.; and Dragan, A.~D. 2019.
\newblock On the utility of model learning in HRI.
\newblock In \emph{International Conference on Human-Robot Interaction}, 317--325. IEEE.

\bibitem[{Chromik, Carlmeyer, and Wrede(2017)}]{chromik2017incremental}
Chromik, M.; Carlmeyer, B.; and Wrede, B. 2017.
\newblock Ready for the Next Step? Investigating the Effect of Incremental Information Presentation in an Object Fetching Task.
\newblock In \emph{Companion of the ACM/IEEE International Conference on Human-Robot Interaction}, 95–96. Association for Computing Machinery.
\newblock ISBN 9781450348850.

\bibitem[{Das et~al.(2019)Das, Gervet, Romoff, Batra, Parikh, Rabbat, and Pineau}]{das2019tarmac}
Das, A.; Gervet, T.; Romoff, J.; Batra, D.; Parikh, D.; Rabbat, M.; and Pineau, J. 2019.
\newblock {TarMAC}: Targeted multi-agent communication.
\newblock In \emph{International Conference on Machine Learning}, 1538--1546. PMLR.

\bibitem[{DeVault, Sagae, and Traum(2009)}]{devault2009can}
DeVault, D.; Sagae, K.; and Traum, D. 2009.
\newblock Can I finish? Learning when to respond to incremental interpretation results in interactive dialogue.
\newblock In \emph{SIGDIAL}, 11--20.

\bibitem[{DeVault, Sagae, and Traum(2011)}]{devault2011incremental}
DeVault, D.; Sagae, K.; and Traum, D. 2011.
\newblock Incremental interpretation and prediction of utterance meaning for interactive dialogue.
\newblock \emph{Dialogue \& Discourse}, 2(1): 143--170.

\bibitem[{Dro{\'z}dziel et~al.(2020)Dro{\'z}dziel, Tarkowski, Rybicka, and Wrona}]{drozdziel2020drivers}
Dro{\'z}dziel, P.; Tarkowski, S.; Rybicka, I.; and Wrona, R. 2020.
\newblock Drivers’ reaction time research in the conditions in the real traffic.
\newblock \emph{Open Engineering}, 10(1): 35--47.

\bibitem[{Fudenberg et~al.(2020)Fudenberg, Newey, Strack, and Strzalecki}]{fudenberg2020testing}
Fudenberg, D.; Newey, W.; Strack, P.; and Strzalecki, T. 2020.
\newblock Testing the drift-diffusion model.
\newblock \emph{Proceedings of the National Academy of Sciences}, 117(52): 33141--33148.

\bibitem[{Gilardi, Alizadeh, and Kubli(2023{\natexlab{a}})}]{gilardi2023chatgpt}
Gilardi, F.; Alizadeh, M.; and Kubli, M. 2023{\natexlab{a}}.
\newblock ChatGPT outperforms crowd workers for text-annotation tasks.
\newblock \emph{Proceedings of the National Academy of Sciences}, 120(30): e2305016120.

\bibitem[{Gilardi, Alizadeh, and Kubli(2023{\natexlab{b}})}]{chatgptannotator}
Gilardi, F.; Alizadeh, M.; and Kubli, M. 2023{\natexlab{b}}.
\newblock ChatGPT outperforms crowd workers for text-annotation tasks.
\newblock \emph{Proceedings of the National Academy of Sciences}, 120(30): e2305016120.

\bibitem[{Gollwitzer(1999)}]{gollwitzer1999implementation}
Gollwitzer, P.~M. 1999.
\newblock Implementation intentions: strong effects of simple plans.
\newblock \emph{American psychologist}, 54(7): 493.

\bibitem[{Hoffman, Bhattacharjee, and Nikolaidis(2023)}]{hoffman2023inferring}
Hoffman, G.; Bhattacharjee, T.; and Nikolaidis, S. 2023.
\newblock Inferring human intent and predicting human action in human--robot collaboration.
\newblock \emph{Annual Review of Control, Robotics, and Autonomous Systems}, 7.

\bibitem[{Hoffman and Breazeal(2004)}]{hoffman2004collaboration}
Hoffman, G.; and Breazeal, C. 2004.
\newblock Collaboration in human-robot teams.
\newblock In \emph{AIAA 1st intelligent systems technical conference}, 6434.

\bibitem[{Hsu et~al.(2025)Hsu, Defranco, Patel, and Nikolaidis}]{hsu2025integrating}
Hsu, Y.-C.; Defranco, M.; Patel, R.; and Nikolaidis, S. 2025.
\newblock Integrating Field of View in Human-Aware Collaborative Planning.
\newblock In \emph{International Conference on Robotics and Automation}.

\bibitem[{Jin and Orosz(2016)}]{jin2016optimal}
Jin, I.~G.; and Orosz, G. 2016.
\newblock Optimal control of connected vehicle systems with communication delay and driver reaction time.
\newblock \emph{IEEE Transactions on Intelligent Transportation Systems}, 18(8): 2056--2070.

\bibitem[{Kesting, Treiber, and Helbing(2007)}]{kesting2007general}
Kesting, A.; Treiber, M.; and Helbing, D. 2007.
\newblock General lane-changing model MOBIL for car-following models.
\newblock In \emph{Transportation Research Board 86th Annual Meeting}.

\bibitem[{Klissarov et~al.(2023)Klissarov, D'Oro, Sodhani, Raileanu, Bacon, Vincent, Zhang, and Henaff}]{klissarov2023motif}
Klissarov, M.; D'Oro, P.; Sodhani, S.; Raileanu, R.; Bacon, P.-L.; Vincent, P.; Zhang, A.; and Henaff, M. 2023.
\newblock Motif: Intrinsic motivation from artificial intelligence feedback.
\newblock \emph{arXiv preprint arXiv:2310.00166}.

\bibitem[{Lee et~al.(2023)Lee, Krishna, Zaidi, Paleja, Chen, Hedlund-Botti, Schrum, and Gombolay}]{lee2023effect}
Lee, K.~M.; Krishna, A.; Zaidi, Z.; Paleja, R.; Chen, L.; Hedlund-Botti, E.; Schrum, M.; and Gombolay, M. 2023.
\newblock The Effect of Robot Skill Level and Communication in Rapid, Proximate Human-Robot Collaboration.
\newblock In \emph{International Conference on Human-Robot Interaction}.

\bibitem[{Leurent(2018)}]{highway-env}
Leurent, E. 2018.
\newblock An Environment for Autonomous Driving Decision-Making.
\newblock \url{https://github.com/eleurent/highway-env}.

\bibitem[{Li et~al.(2024)Li, Xu, Sachdeva, Misu, and Dariush}]{li2024optimal}
Li, C.; Xu, A.; Sachdeva, E.; Misu, T.; and Dariush, B. 2024.
\newblock Optimal Driver Warning Generation in Dynamic Driving Environment.
\newblock In \emph{International Conference on Robotics and Automation}, 14184--14190. IEEE.

\bibitem[{Liu et~al.(2024)Liu, Yu, Gao, Xie, Liao, Wu, and Wang}]{liu2024llm}
Liu, J.; Yu, C.; Gao, J.; Xie, Y.; Liao, Q.; Wu, Y.; and Wang, Y. 2024.
\newblock {LLM}-powered hierarchical language agent for real-time human-ai coordination.
\newblock \emph{Proceedings of the 23rd International Conference on Autonomous Agents and Multiagent Systems (AAMAS)}.

\bibitem[{Lopes~Rego, Snell, and Meeter(2024)}]{lopes2024language}
Lopes~Rego, A.~T.; Snell, J.; and Meeter, M. 2024.
\newblock Language models outperform cloze predictability in a cognitive model of reading.
\newblock \emph{PLOS Computational Biology}, 20(9): e1012117.

\bibitem[{Luebbers et~al.(2023)Luebbers, Tabrez, Ruvane, and Hayes}]{luebbers2023autonomous}
Luebbers, M.~B.; Tabrez, A.; Ruvane, K.; and Hayes, B. 2023.
\newblock Autonomous Justification for Enabling Explainable Decision Support in Human-Robot Teaming.
\newblock In \emph{Robotics: Science and Systems}.

\bibitem[{Mandi, Jain, and Song(2024)}]{mandi2024roco}
Mandi, Z.; Jain, S.; and Song, S. 2024.
\newblock Roco: Dialectic multi-robot collaboration with large language models.
\newblock In \emph{International Conference on Robotics and Automation}, 286--299. IEEE.

\bibitem[{Mnih et~al.(2013)Mnih, Kavukcuoglu, Silver, Graves, Antonoglou, Wierstra, and Riedmiller}]{mnih2013playing}
Mnih, V.; Kavukcuoglu, K.; Silver, D.; Graves, A.; Antonoglou, I.; Wierstra, D.; and Riedmiller, M. 2013.
\newblock Playing {A}tari with deep reinforcement learning.
\newblock \emph{arXiv preprint arXiv:1312.5602}.

\bibitem[{Mohammad, Farah, and Zgonnikov(2024)}]{mohammad2024driver}
Mohammad, S.~H.; Farah, H.; and Zgonnikov, A. 2024.
\newblock In the driver's mind: Modeling the dynamics of human overtaking decisions in interactions with oncoming automated vehicles.
\newblock \emph{Transportation Research Part F: Traffic Psychology and Behaviour}, 107: 562--577.

\bibitem[{Nikolaidis et~al.(2018)Nikolaidis, Kwon, Forlizzi, and Srinivasa}]{nikolaidis2018planning}
Nikolaidis, S.; Kwon, M.; Forlizzi, J.; and Srinivasa, S. 2018.
\newblock Planning with verbal communication for human-robot collaboration.
\newblock \emph{ACM Transactions on Human-Robot Interaction}, 7(3): 1--21.

\bibitem[{Park et~al.(2023)Park, O'Brien, Cai, Morris, Liang, and Bernstein}]{park2023generative}
Park, J.~S.; O'Brien, J.; Cai, C.~J.; Morris, M.~R.; Liang, P.; and Bernstein, M.~S. 2023.
\newblock Generative agents: Interactive simulacra of human behavior.
\newblock In \emph{{ACM} symposium on user interface software and technology}, 1--22.

\bibitem[{Ratcliff, Huang-Pollock, and McKoon(2018)}]{ratcliff2018modeling}
Ratcliff, R.; Huang-Pollock, C.; and McKoon, G. 2018.
\newblock Modeling individual differences in the go/no-go task with a diffusion model.
\newblock \emph{Decision}, 5(1): 42.

\bibitem[{Reimann et~al.(2024)Reimann, Kunneman, Oertel, and Hindriks}]{reimann2024survey}
Reimann, M.~M.; Kunneman, F.~A.; Oertel, C.; and Hindriks, K.~V. 2024.
\newblock A survey on dialogue management in human-robot interaction.
\newblock \emph{ACM Transactions on Human-Robot Interaction}, 13(2): 1--22.

\bibitem[{Schulman et~al.(2017)Schulman, Wolski, Dhariwal, Radford, and Klimov}]{schulman2017proximal}
Schulman, J.; Wolski, F.; Dhariwal, P.; Radford, A.; and Klimov, O. 2017.
\newblock Proximal policy optimization algorithms.
\newblock \emph{arXiv preprint arXiv:1707.06347}.

\bibitem[{Shvo et~al.(2022)Shvo, Hari, O'Reilly, Abolore, Wang, and McIlraith}]{shvo2023proactive}
Shvo, M.; Hari, R.; O'Reilly, Z.; Abolore, S.; Wang, S.-Y.~N.; and McIlraith, S.~A. 2022.
\newblock Proactive Robotic Assistance via Theory of Mind.
\newblock In \emph{International Conference on Intelligent Robots and Systems (IROS)}, 9148--9155.

\bibitem[{Shvo, Klassen, and McIlraith(2022)}]{shvo2022resolving}
Shvo, M.; Klassen, T.~Q.; and McIlraith, S.~A. 2022.
\newblock Resolving misconceptions about the plans of agents via Theory of Mind.
\newblock In \emph{Proceedings of the International Conference on Automated Planning and Scheduling}, volume~32, 719--729.

\bibitem[{Srikanth et~al.(2025)Srikanth, Bhatt, Zhang, Hager, Lewis, Sycara, Tabrez, and Nikolaidis}]{srikanth2025algorithmic}
Srikanth, S.; Bhatt, V.; Zhang, B.; Hager, W.; Lewis, C.~M.; Sycara, K.~P.; Tabrez, A.; and Nikolaidis, S. 2025.
\newblock Algorithmic Prompt Generation for Diverse Human-like Teaming and Communication with Large Language Models.
\newblock \emph{arXiv preprint arXiv:2504.03991}.

\bibitem[{St.~Clair and Mataric(2015)}]{st2015robot}
St.~Clair, A.; and Mataric, M. 2015.
\newblock How robot verbal feedback can improve team performance in human-robot task collaborations.
\newblock In \emph{International Conference on Human-Robot Interaction}, 213--220.

\bibitem[{Sundareswara et~al.(2013)Sundareswara, Daily, Howard, Neely, Bhattacharyya, and Lee}]{sundareswara2013using}
Sundareswara, R.; Daily, M.; Howard, M.; Neely, H.; Bhattacharyya, R.; and Lee, C. 2013.
\newblock Using a distracted driver's behavior to inform the timing of alerts in a semi-autonomous car.
\newblock In \emph{International Multi-Disciplinary Conference on Cognitive Methods in Situation Awareness and Decision Support}, 199--202. IEEE.

\bibitem[{Sutton, Barto et~al.(1998)}]{sutton1998reinforcement}
Sutton, R.~S.; Barto, A.~G.; et~al. 1998.
\newblock \emph{Reinforcement learning: An introduction}, volume~1.
\newblock MIT press Cambridge.

\bibitem[{Treiber, Hennecke, and Helbing(2000)}]{treiber2000congested}
Treiber, M.; Hennecke, A.; and Helbing, D. 2000.
\newblock Congested traffic states in empirical observations and microscopic simulations.
\newblock \emph{Transportation Science}, 39(2): 147--159.

\bibitem[{Trueblood et~al.(2011)Trueblood, Endres, Busemeyer, and Finn}]{trueblood2011modeling}
Trueblood, J.~S.; Endres, M.~J.; Busemeyer, J.~R.; and Finn, P.~R. 2011.
\newblock Modeling response times in the go/no-go discrimination task.
\newblock In \emph{Cogsci}, volume 2011, 1866.

\bibitem[{Tucker et~al.(2025)Tucker, Shah, Levy, and Zaslavsky}]{10.1162/opmi_a_00188}
Tucker, M.; Shah, J.; Levy, R.; and Zaslavsky, N. 2025.
\newblock Towards Human-Like Emergent Communication via Utility, Informativeness, and Complexity.
\newblock \emph{Open Mind}, 9: 418--451.

\bibitem[{Unhelkar and Shah(2016)}]{unhelkar2016contact}
Unhelkar, V.; and Shah, J. 2016.
\newblock Contact: Deciding to communicate during time-critical collaborative tasks in unknown, deterministic domains.
\newblock In \emph{AAAI Conference on Artificial Intelligence}, volume~30.

\bibitem[{Unhelkar, Li, and Shah(2020)}]{unhelkar2020decision}
Unhelkar, V.~V.; Li, S.; and Shah, J.~A. 2020.
\newblock Decision-making for bidirectional communication in sequential human-robot collaborative tasks.
\newblock In \emph{International Conference on Human-Robot Interaction}, 329--341.

\bibitem[{Wang et~al.(2024)Wang, Pant, Zhao, Antkiewicz, and Czarnecki}]{wang2024enhancing}
Wang, J.; Pant, Y.~V.; Zhao, L.; Antkiewicz, M.; and Czarnecki, K. 2024.
\newblock Enhancing safety in mixed traffic: Learning-based modeling and efficient control of autonomous and human-driven vehicles.
\newblock \emph{IEEE Transactions on Intelligent Transportation Systems}.

\bibitem[{Wang et~al.(2022)Wang, Zhong, Xu, and Wang}]{wang2021tom2c}
Wang, Y.; Zhong, F.; Xu, J.; and Wang, Y. 2022.
\newblock Tom2c: Target-oriented multi-agent communication and cooperation with theory of mind.
\newblock \emph{ICLR}.

\bibitem[{Yuan, Liberman, and Cieri(2006)}]{yuan2006towards}
Yuan, J.; Liberman, M.; and Cieri, C. 2006.
\newblock Towards an integrated understanding of speaking rate in conversation.
\newblock In \emph{International Conference on Spoken Language Processing}.

\bibitem[{Zhang et~al.(2024{\natexlab{a}})Zhang, Yang, Hu, Wang, Li, Sun, Zhang, Zhang, Liu, Zhu et~al.}]{zhang2024proagent}
Zhang, C.; Yang, K.; Hu, S.; Wang, Z.; Li, G.; Sun, Y.; Zhang, C.; Zhang, Z.; Liu, A.; Zhu, S.-C.; et~al. 2024{\natexlab{a}}.
\newblock Proagent: building proactive cooperative agents with large language models.
\newblock In \emph{AAAI Conference on Artificial Intelligence}, volume~38, 17591--17599.

\bibitem[{Zhang et~al.(2024{\natexlab{b}})Zhang, Wang, Gong, Lin, and Mao}]{zhang2024usimagent}
Zhang, E.; Wang, X.; Gong, P.; Lin, Y.; and Mao, J. 2024{\natexlab{b}}.
\newblock Usimagent: Large language models for simulating search users.
\newblock In \emph{Proceedings of the 47th International ACM SIGIR Conference on Research and Development in Information Retrieval}, 2687--2692.

\bibitem[{Zhang et~al.(2024{\natexlab{c}})Zhang, Du, Shan, Zhou, Du, Tenenbaum, Shu, and Gan}]{zhang2023building}
Zhang, H.; Du, W.; Shan, J.; Zhou, Q.; Du, Y.; Tenenbaum, J.~B.; Shu, T.; and Gan, C. 2024{\natexlab{c}}.
\newblock Building cooperative embodied agents modularly with large language models.
\newblock \emph{ICLR}.

\bibitem[{Zhao et~al.(2021)Zhao, Fan, Li, Zheng, and Pan}]{zhao2021efficient}
Zhao, X.; Fan, T.; Li, Y.; Zheng, Y.; and Pan, J. 2021.
\newblock An efficient and responsive robot motion controller for safe human-robot collaboration.
\newblock \emph{IEEE Robotics and Automation Letters}, 6(3): 6068--6075.

\bibitem[{Zheng et~al.(2023)Zheng, Chiang, Sheng, Zhuang, Wu, Zhuang, Lin, Li, Li, Xing et~al.}]{zheng2023judging}
Zheng, L.; Chiang, W.-L.; Sheng, Y.; Zhuang, S.; Wu, Z.; Zhuang, Y.; Lin, Z.; Li, Z.; Li, D.; Xing, E.; et~al. 2023.
\newblock Judging llm-as-a-judge with mt-bench and chatbot arena.
\newblock \emph{Advances in Neural Information Processing Systems}, 36: 46595--46623.

\end{thebibliography}
\clearpage

\appendix

\section{Informative content types}
\label{app:informative_content_types}
Throughout our experiments, we adopt an operational definition of informativeness that is proportional to utterance length ($I(\mathbf{u}) \propto l$) and a timing parameterization via the comprehension point $k$. This abstraction is intentional: it allows us to analyze timeliness–informativeness trade-offs independent of surface form. For completeness, we provide a semantic instantiation that decomposes utterances into \emph{What} (action), \emph{How} (procedure/parameters), and \emph{Why} (rationale), whose ordering induces the same $(k,l)$ and can optionally contribute to informativeness. Here, $k$ is the position where \emph{What} first appears and $l$ is the total length. We assume each notification contains exactly one action directive (What), such that an actionable $k$ is defined. For example,
\begin{itemize}
    \item $\mathbf{u_1}$: ``Slowly press the pedal to speed up." ($k=1$; How-What-Why)
    \item $\mathbf{u_2}$: ``Press the pedal slowly to speed up." ($k=0$; What-How-Why)
    \item $\mathbf{u_3}$: ``To speed up, press the pedal slowly." ($k=3$; Why-What-How)
    \item $\mathbf{u_4}$: ``You are below the minimal limit. To speed up, press the pedal slowly." ($k=9$; Why-What-How)
\end{itemize}
This grounds $(k, l)$ in linguistic design without changing algorithms or metrics. When desired, informativeness can be enriched as $I(\mathbf{u}) = \alpha l + \beta I_{why}$, where $I_{why}$ captures the contribution of rationale (Why) content. In the examples above, $I_{why}(\mathbf{u}_1) = 0$, $I_{why}(\mathbf{u}_2) = 0$, $I_{why}(\mathbf{u}_3) > 0$, and $I_{why}(\mathbf{u}_4) > I_{why}(\mathbf{u}_3) > 0$. In the main paper, we set $\beta = 0$ to match the operational assumption $I(\mathbf{u}) \propto l$.

In time-critical phases, we prefer $k=0$ (What–How–Why) to ensure immediate actionability. When slack exists, we allow $k \le k_{\max}$—an upper bound on delay before revealing the action—learned end-to-end during training. This permits Why/How to precede What: brief rationales can strengthen credibility and calibrate trust, while early procedural qualifiers can improve action precision (e.g., with delicate objects). Our framework can accommodate these cases by enriching the informativeness measure and its domain-specific effects—i.e., by adjusting $I(\mathbf{u})$ and the mapping from $I(\mathbf{u})$ to reaction reward and follow-through behavior.

\section{Domain taxonomy}
\label{app:domain_taxonomy}
We define the domain taxonomy for our offline–generated utterance database.  
Each entry is a triple $(c,k,l)$, where:
\begin{itemize}
  \item $c$ is a content topic (abstract instruction);
  \item $k$ is the \emph{actionable index} (word position at which the human can begin to react);
  \item $l$ is the total length in words of the utterance.
\end{itemize}
Table~\ref{app:tab:tax_topics} lists the set of content topics $C$ for each domain. The values for $k$ and $l$ are $[2,5]$ for all three domains. We sample a handful of example triples and their corresponding English utterances in Table~\ref{app:tab:tax_examples}. The entire dataset is shared in the source code.

\begin{table}[ht]
  \centering
  \resizebox{\linewidth}{!}{%
  \begin{tabular}{@{}l l@{}}
    \toprule
    \textbf{Domain} & \textbf{Content Topics $C$} \\
    \midrule
    Piloting (Lunar Lander) & ascend, shift right/left, descend \\
    Driving  (Highway Merge) & slow down, speed up, merge left/right \\
    Cooking  (Steakhouse)    & move up/down/left/right, disclose information \\
    \bottomrule
  \end{tabular}
  }
  \caption{Domain‐specific content topics}
  \label{app:tab:tax_topics}
\end{table}

\begin{table}[ht]
  \centering
  \footnotesize
  \begin{tabular}{@{}l l c c l@{}}
    \toprule
    \textbf{Domain} & $c$      & $k$ & $l$ & \textbf{Example Utterance} \\ 
    \midrule
    Piloting        & left     & 2   & 2   & “Shift left.” \\
    Piloting        & ascend    & 5   & 5   & “Avoid top-left danger zone, ascend.” \\
    Driving         & slow     & 2   & 5   & “Slow down for incoming traffic.” \\
    Driving         & right     & 5   & 5   & “Merge to the right lane.” \\
    Cooking         & down     & 2   & 2   & “Go down.” \\
    Cooking         & right   & 5   & 5   & “Chopping board full, go right.” \\
    \bottomrule
  \end{tabular}
  \caption{Sample taxonomy entries $(c,k,l)$ with utterances}
  \label{app:tab:tax_examples}
\end{table}

\section{Domain design for notifier assistive setting}
\label{app:domain_details}
We introduce modifications to the Lunar Lander, Highway Merge, and Steakhouse environments to model scenarios where a human operator critically depends on an automated notifier due to limited situational awareness. 

In each domain, the notifier holds essential information inaccessible to the human agent. In Lunar Lander, we introduced invisible danger zones simulating environmental hazards such as turbulence or overlapping flight paths. These danger zones are detectable only by a central control notifier, thereby making timely and informative communication crucial to safe navigation. In Highway Merge, we design three merging events as key decision points. Merging vehicles do not detect the ego vehicle, whereas the ego vehicle driver operates under the assumption that merging vehicles will yield. This mismatch in expectations creates frequent collision risks. Additionally, one of these merge lanes approaches from the left instead of the traditional right-hand side, increasing unpredictability. Finally, in the Steakhouse setting, we adjusted the initial conditions by pre-occupying kitchen stations, introducing uncertainty in the human agent's available workspace. Additionally, the agent's visual awareness was constrained to a limited 2-by-3 grid directly ahead, simulating realistic perceptual limitations. However, this results in a higher probability of performing redundant actions since the agent will only know the occupancy of a station once near arrival.

\section{Human base policies}
\label{app:human_agent_polices}
In this work, we evaluate notification policies against an utterance-reactive human model \(H_{react}\) that is capable of decision making for completing the task, formulated as a Markov Decision Process, denoted as MDP, and also reacts to provide utterances via the $\mathcal{M}_{react}$ model. In this section, we detail MDP, designed to reflect realistic, domain-specific behavior:

\begin{itemize}
  \item \textbf{Lunar Lander.}  
    Here, MDP is a deep‐RL policy trained in the OpenAI Gymnasium LunarLander environment~\cite{brockman2016openai}. We use the default observation space, action space, and reward function. To ensure that the policy remains robust when notifications perturb the state—and does not stray outside its training distribution—we train it with the environment’s turbulence and wind settings enabled.
    
  \item \textbf{Highway Merge.}  
    In this domain, the task-completion MDP is instantiated as a rule-based policy using the Intelligent Driver Model (IDM)~\cite{treiber2000congested}, augmented with the MOBIL lane-change criterion~\cite{kesting2007general}. IDM defines deterministic state transitions for longitudinal acceleration as a function of vehicle spacing and relative speed, while MOBIL governs lateral transitions by evaluating acceleration impacts on nearby vehicles. Although no explicit reward function is optimized, this rule-base policy can be interpreted under our MDP definition as encoding implicit rewards for maintaining safety and efficiency, with transitions determined by the IDM/MOBIL.
    
  \item \textbf{Steakhouse.}  
    The MDP is a myopic, rule-based agent inspired by Overcooked-AI benchmarks~\cite{hsu2025integrating, carroll2019utility}. At each step, it selects the highest-priority available subtask --- such as picking up meat when the grill is empty or fetching a missing ingredient. When multiple subtasks share the same priority, it breaks ties uniformly at random. This simple heuristic captures a human who focuses on immediate progress without long-term planning.
\end{itemize}

By fixing MDP in each domain, we establish a consistent baseline against which to measure how our timing- and content-aware notification policies influence task success and safety.

\section{Notifier architecture}
\label{app:notifier_architecture}                              
Our notifier architecture extends the standard multilayer perceptron (MLP) setup (replaced by a small LSTM in the Steakhouse domain to better handle partial observability) with domain-adaptive input processing and a multi-head output layer to jointly predict all aspects of the notification. At each time step, we concatenate the current observation with previous actions and pass through two 64-unit fully connected layers with $\tanh$ activations. From this, we then output up to four linear heads, each parameterizing a categorical distribution over one component of the notification tuple. A separate critic network operates on the same processed input to estimate state values. We train the entire architecture end-to-end with Proximal Policy Optimization (PPO), summing log-probabilities and entropies across policy heads to form a standard PPO objective. This design cleanly decouples domain-specific feature encoding from our general MLP (and LSTM) backbone and allows flexible configuration of notification parameters while ensuring stable learning.

Domain-specific observation processing is described as follows:
\begin{itemize}
  \item \textbf{Lunar Lander.} The observation is an 8-dimensional vector. We process it by concatenating the current observation with past observations and actions to form the input to the MLP backbone.
    
  \item \textbf{Highway Merge.} We adopt the transformer encoder architecture from the open-source highway domain codebase~\cite{highway-env}. Each vehicle is embedded individually, and the set of vehicle embeddings is passed through an ego-centered multi-head attention network, where the ego vehicle is used as queries and all vehicles serve as keys and values. The resulting attention-based feature representation is then concatenated with past observation features and actions to create the input to the MLP backbone.
    
  \item \textbf{Steakhouse.} We extract only the observations relevant to notifier decision-making: the human's position and orientation, the states of key objects (e.g., progress on washing dirty plates), and a boolean flag indicating whether the human is aware of each object’s state. At each time step, the current observation is concatenated with the past observations and actions and fed into an LSTM notifier backbone, which also outputs four linear heads as in the MLP backbone. Detailed LSTM parameters are shown in Table~\ref{tab:arch_details}.
\end{itemize}

We present hyperparameters for PPO in Table \ref{tab:ppo_details} and model architectures in Table \ref{tab:arch_details}. 
All experiments were conducted on four compute clusters, each with a single NVIDIA RTX Ada 6000 48 GB GPU.

\begin{table}[ht]
  \centering
  \footnotesize
  \setlength{\tabcolsep}{6pt}  
  \renewcommand{\arraystretch}{0.9}
  \begin{tabular}{ll|ll}
    \toprule
    \textbf{Hyperparameter} & \textbf{Value} 
      & \textbf{Hyperparameter} & \textbf{Value} \\
    \midrule
    \texttt{learning\_rate}   & 2.5e-4  & \texttt{anneal\_lr}        & True   \\
    \texttt{num\_envs}        & 4       & \texttt{gamma}             & 0.99   \\
    \texttt{num\_steps}       & 128     & \texttt{gae\_lambda}       & 0.95   \\
    \texttt{num\_minibatches} & 4       & \texttt{update\_epochs}    & 4      \\
    \texttt{norm\_adv}        & True    & \texttt{clip\_coef}        & 0.2    \\
    \texttt{clip\_vloss}      & True    & \texttt{ent\_coef}         & 0.01   \\
    \texttt{vf\_coef}         & 0.5     & \texttt{max\_grad\_norm}   & 0.5    \\
    \texttt{target\_kl}       & None    &                            &        \\
    \bottomrule
  \end{tabular}
  \caption{PPO Settings}
  \label{tab:ppo_details}
\end{table}

\begin{table}[ht]
\centering
\footnotesize
\begin{tabular}{ll}
\toprule
\textbf{Hyperparameter} & \textbf{Value} \\
\midrule
\textbf{MLP} & \\
\texttt{mlp\_hidden\_dims} & [64, 64] \\
\textbf{LSTM} & \\
\texttt{lstm\_size} & 32 \\
\texttt{lstm\_hidden\_dim} & 128 \\
\texttt{lstm\_num\_layers} & 1 \\
\bottomrule
\end{tabular}
\caption{Model Architecture Settings}
\label{tab:arch_details}
\end{table}

\begin{figure*}[ht]
    \centering
    \includegraphics[width=0.9\textwidth]{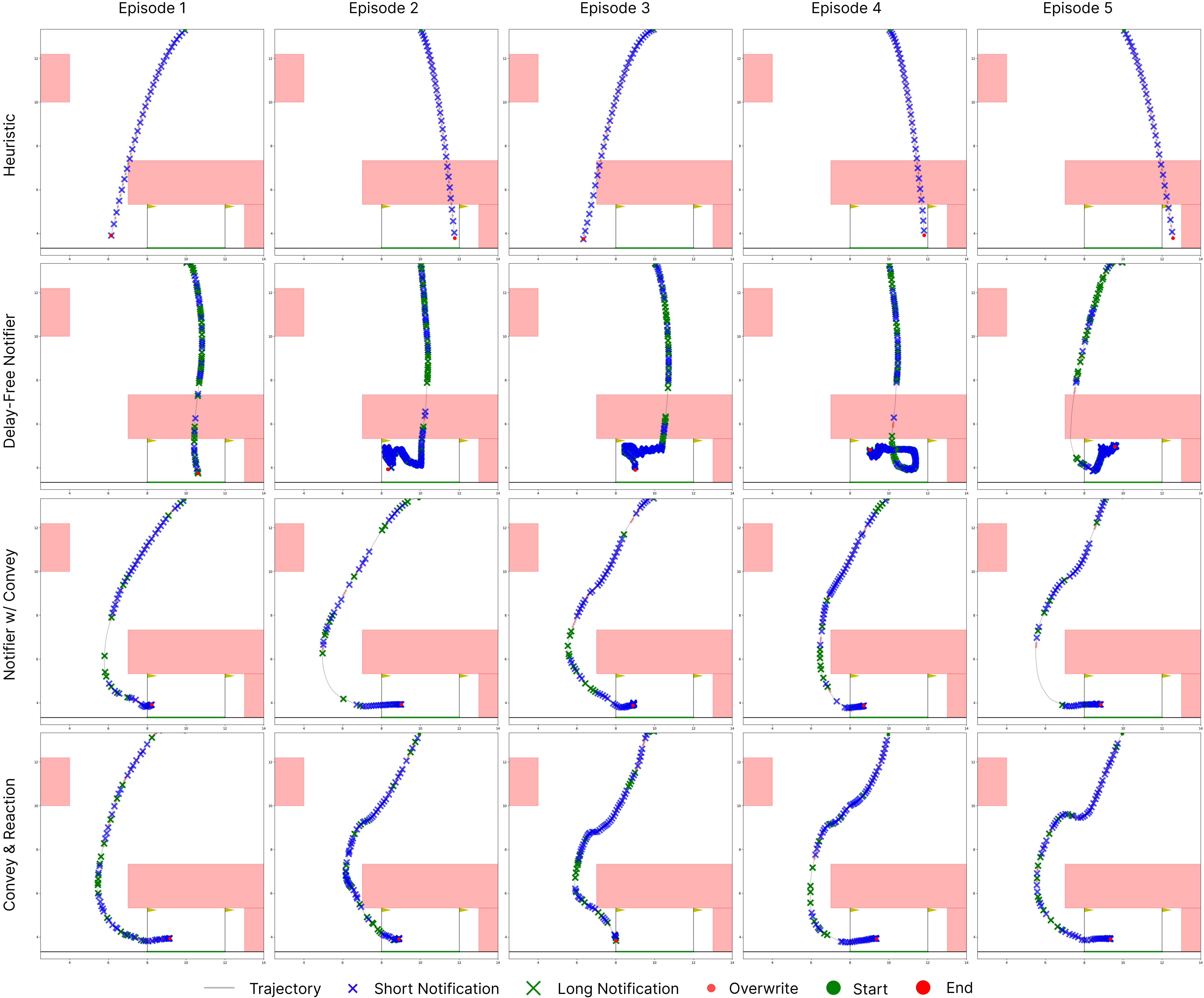}
    \caption{Play through with different policies with the same environment seed. The top row is with a heuristic notifier. The second row is a delay-free notifier. The third row has a notifier that is conveyance time aware. The last row is a notifier with awareness of conveyance and human reaction delay times.}
\label{abs:notifier_policies:ll_play_through}
\end{figure*}

\section{Additional policy behavior analysis}
\label{app:notifier_polices}
\subsection{Statistical significance}
We report the statistical significance of the results in the main paper. Each policy was trained with 5 different seeds and evaluated over 100 rollout episodes per seed. The collected rollouts were then analyzed, and the results were aggregated across all seeds. 

\subsection{Lunar lander notifier policies}
\label{app:ll_policies}
In this section, we describe the implementation of the Heuristic Notifier, the reward designed for the domain, and visualize the different notifier-guided lunar lander trajectories.

\subsubsection{Heuristic notifier}
The heuristic notifier notifies the human when a certain measurement exceeds a threshold, indicating that the human would encounter a dangerous situation without notification. The given notification contains instructions to guide the human onto an optimal path precomputed by the heuristic notifier.

\subsubsection{Reward design}
We show the designed rewards for training the notifier policy operating in the lunar lander domain in Table~\ref{app:tab:lunar_rewards}.

\begin{table}[ht]
\centering
\resizebox{\linewidth}{!}{%
\begin{tabular}{l|p{4cm}|c}
\toprule
\textbf{Reward Name} & \textbf{Description/Condition} & \textbf{Reward Value} \\
\midrule
Fuel usage & Fuel consumed when an engine is active & $-0.3$ (main), $-0.03$ (side) \\
\midrule
Notification initiation & Notifier initiates a new notification & $-1$ \\
\midrule
Near danger zone & Lunar lander within 0.2 distance from danger zones (top, below, left, right); applies if distance $< 0.2$ & $-10 \times$ sum of distances \\
\midrule
In danger zone & Lunar lander enters the danger zone & $-20$ \\
\midrule
Crash landing & Lunar lander crashes to the ground & $-300$ \\
\midrule
Success landing & Lunar lander successfully stationary on the landing pad & $300$ \\
\bottomrule
\end{tabular}
}
\caption{Rewards for Lunar Lander Notifier policy training}
\label{app:tab:lunar_rewards}
\end{table}

\subsubsection{Follow-through duration given informativeness}
\label{app:ll_informativeness}
In this domain, the follow-through duration $d_f$ depends linearly on utterance informativeness $I(\mathbf{u})$. We instantiate $I(\mathbf{u})$ as the message length $l$ (in words), hence $d_f(I(\mathbf{u})) = I(\mathbf{u}) = l$.

\subsubsection{Notifier-guided pilot trajectories}
The trajectories in Fig.~\ref{abs:notifier_policies:ll_play_through} highlight clear different behaviors of each notifier policy. The heuristic notifier consistently fails to guide the pilot away from danger zones, as indicated by straight-line trajectories passing directly through hazardous regions. Similarly, the Delay-Free notifier (RL-trained policy that does not consider time delays) frequently issues ineffective notifications, resulting in erratic behavior and poor performance. In contrast, notifier policies that explicitly consider conveyance duration, particularly our method that factors in human reaction time, produce smoother and safer trajectories. The visualized paths clearly illustrate timely and well-paced notifications, denoted by gray points between notifications (blue dots) and follow-throughs (red dots). Among these policies, explicitly modeling both conveyance and reaction times yields the most accurate and efficient navigation.

\begin{figure*}[ht]
    \centering
    \includegraphics[width=\textwidth]{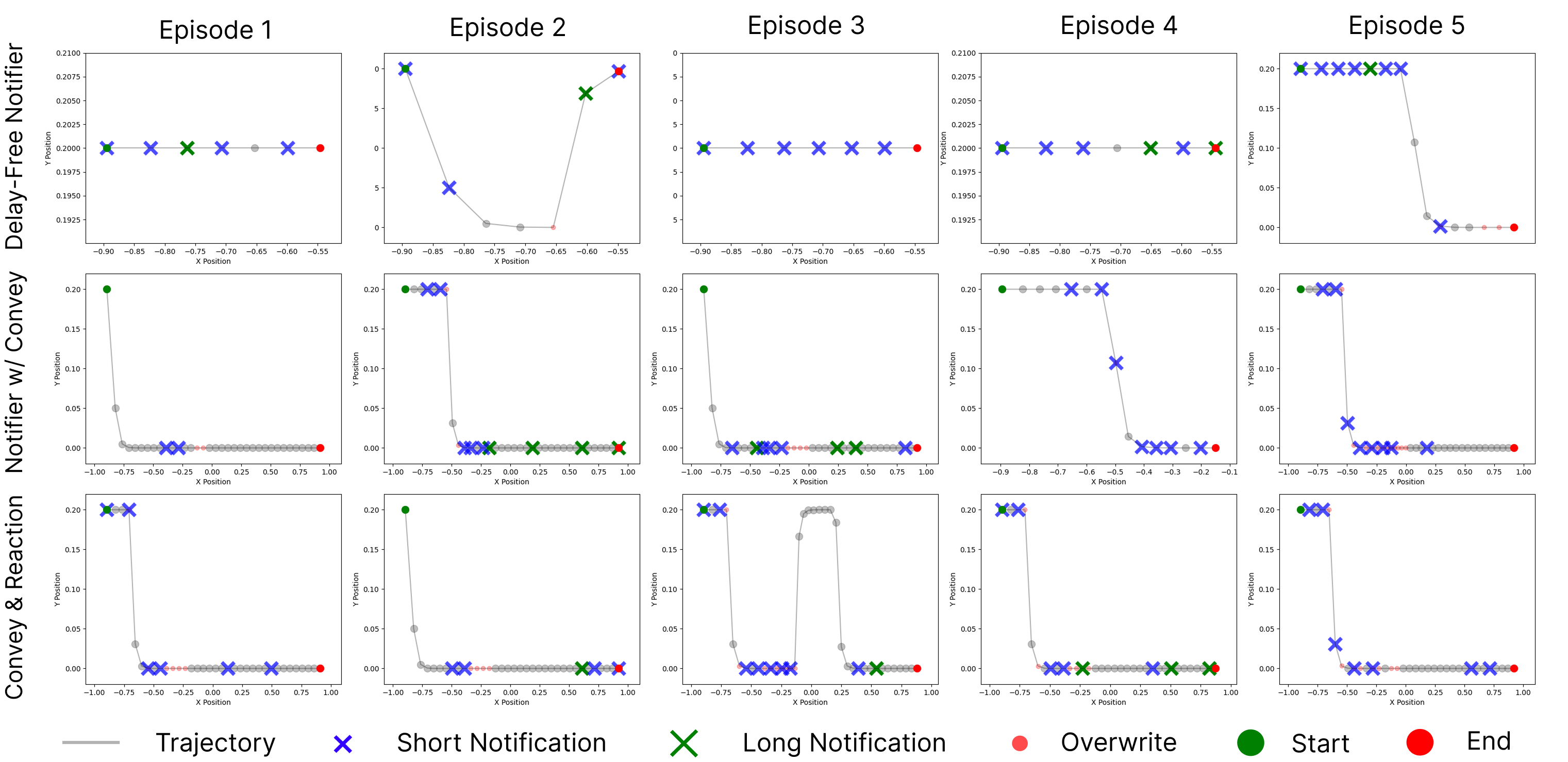}
    \caption{Play through with different policies with the same environment seed in the Driving domain. The changes in the Y-position indicate a lane change. The top row is a Delay-Free notifier. The second row has a notifier that is conveyance time aware. The last row is a notifier aware of conveyance and human reaction delay times.}
\label{abs:notifier_policies:highway_play_through}
\end{figure*}

\subsection{Highway merging notifier policies}
We describe the changes introduced to the original highway merge domain rewards for the notifier policy training and analyze the driving behavior of the Highway notifier guided driver.

\subsubsection{Reward design}
We follow the highway merge environment reward setting for the notifier policy training while increasing the range of speed reward, as we tend to notify the ego vehicle to slow down or speed up for collision avoidance and travel distance efficiency. Specifically, we set \texttt{reward\_speed\_range} to $[15, 38]$. Additionally, we added a notification initialization reward: $-0.3$.

\subsubsection{Follow-through duration given informativeness}
\label{app:highway_informativeness}
In this domain, follow-through duration scales linearly with informativeness. With $I(\mathbf{u})=l$ (message length, in words), this yields $d_f(I(\mathbf{u})) = 2I(\mathbf{u})-2 = 2l-2$. We assign longer durations in this domain since “slow down” and “speed up” alerts are better suited to extended reaction times.

\subsubsection{Notifier-guided driver trajectories}
In Fig.~\ref{abs:notifier_policies:highway_play_through}, the Delay‐Free notifier has almost no overwritten notifications as the policy assumes instantaneous comprehension. By contrast, the conveyance‐time–aware notifiers (Notifier w/ Convey and Convey \& React) send their messages with appropriate lead time: we observe a higher rate of meaningful follow-through events after each notification, indicating that these policies more effectively guide the driver to slow down or change lanes when needed.

\begin{figure}[ht]
    \centering
    \includegraphics[width=\linewidth]{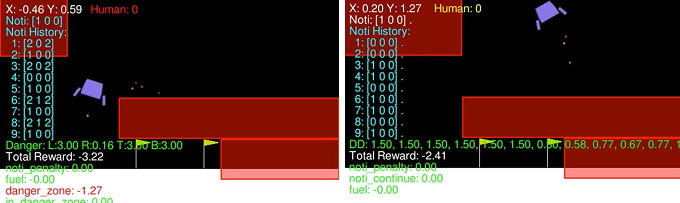}
    \caption{Left: Original Lunar Lander environment; Right: hard version of the Lunar Lander Domain.}
    \label{app:fig:ori_vs_hard_ll}
\end{figure}

\section{Lunar lander domain (Harder version)}
\label{app:ll_hard}
We examine a more challenging variant of the Lunar Lander domain. The difficulty is increased by expanding the top-left danger zone, significantly reducing the gap between the top-left and center danger zones (see Fig.~\ref{app:fig:ori_vs_hard_ll}). This adjustment demands more precise notification control. The motivation behind this change is to further penalize early notifications within the domain. While the success rate drops due to the environment difficulty, our method exceeds other baselines on most metrics (see Table~\ref{app:tab:extended_lunar_lander}).

We present hyperparameters for PPO in Table~\ref{tab:ll_hard_ppo_details}. Additionally, we normalize the observation and reward values for stable training.

\begin{table}[ht]
  \centering
  \footnotesize
  \setlength{\tabcolsep}{6pt}  
  \renewcommand{\arraystretch}{0.9}
  \begin{tabular}{ll|ll}
    \toprule
    \textbf{Hyperparameter} & \textbf{Value} 
      & \textbf{Hyperparameter} & \textbf{Value} \\
    \midrule
    \texttt{learning\_rate}   & 2.5e-4  & \texttt{anneal\_lr}        & True   \\
    \texttt{num\_envs}        & 16       & \texttt{gamma}             & 0.99   \\
    \texttt{num\_steps}       & 600     & \texttt{gae\_lambda}       & 0.95   \\
    \texttt{num\_minibatches} & 8       & \texttt{update\_epochs}    & 8      \\
    \texttt{norm\_adv}        & True    & \texttt{clip\_coef}        & 0.2    \\
    \texttt{clip\_vloss}      & True    & \texttt{ent\_coef}         & 0.01   \\
    \texttt{vf\_coef}         & 0.5     & \texttt{max\_grad\_norm}   & 0.5    \\
    \texttt{target\_kl}       & None    &                            &        \\
    \bottomrule
  \end{tabular}
  \caption{PPO Settings}
  \label{tab:ll_hard_ppo_details}
\end{table}

\begin{table*}[t]
\centering
\resizebox{0.8\textwidth}{!}{%
\begin{tabular}{l|c||cccc}
\toprule
\multirow{2}{*}{\textbf{Notifier Policy}}
  & \multirow{2}{*}{\textbf{Success Rate (↑)}}
    & \multicolumn{4}{c}{\textbf{Secondary Metrics}} \\
\cmidrule(lr){3-6}
  & 
    & \textbf{Noti. Freq} 
    & \textbf{Follow-Through} 
    & \textbf{Long Noti.} 
    & \textbf{Enter Danger} \\
\midrule
Delay-Free Notifier     
    & $0.02 \pm 0.01$      
    & $0.21 \pm 0.06$      
    & \cellcolor{best}$0.28 \pm 0.05$      
    & N/A      
    & $0.98 \pm 0.01$     \\

Notifier w/ Convey      
    & $0.75 \pm 0.41$      
    & $0.70 \pm 0.07$      
    & $0.17 \pm 0.04$      
    & $0.14 \pm 0.23$      
    & $0.18 \pm 0.30$     \\

Convey \& React      
    & \cellcolor{best}$0.83 \pm 0.11$ 
    & \cellcolor{best}$0.72 \pm 0.02$      
    & $0.17 \pm 0.03$      
    & \cellcolor{best}$0.11 \pm 0.09$      
    & \cellcolor{best}$0.06 \pm 0.03$     \\
\bottomrule
\end{tabular}%
}
\caption{Results of guiding pilot through challenging danger zones. Convey \& React (our method) outperforms the baselines in most metrics. Blue highlights the best performance within each column.}
\label{app:tab:extended_lunar_lander}
\end{table*}

\subsubsection{Reward given informativeness}
\label{app:hard_ll_informativenss}
In this domain, we fix the follow-through duration to a constant $d_f = 3$. We instantiate informativeness as the notification length, $I(\mathbf{u}) = l$ (in words), and tie it to the reaction reward via
$R^h_{\mathrm{react}}(I(\mathbf{u})) = I(\mathbf{u}) - 2 = l - 2$.

\subsubsection{Notifier-guided piloting trajectories}
We visualize the trajectories in Fig.~\ref{app:fig:hard_ll_traj}, focusing on Notifier w/ Convey and our method, Convey \& React. Results show that Convey \& React tends to issue notifications (blue dots) earlier in the trajectory, accounting for the time humans need to comprehend and respond. In contrast, Notifier w/ Convey delays notifications, as it does not consider human reaction time and instead seeks to avoid premature alerts that might lead the pilot into the top-left danger zone.

\section{Offline taxonomy generation via LLM}
\label{app:noti_analysis}
In this section, we detail the steps for generating the human reaction-aware offline taxonomy, which include summarizing the domain environment, task, and missing information; generating notifications based on specified properties; rating comprehension progression; and labeling notifications accordingly. We utilize Meta-Llama-3-8B-Instruct for summarization, generation, and comprehension ratings. To ensure diversity, we cluster generated notifications using embeddings from a sentence transformer (Instructor-XL model) and select $k$ medoids for inclusion in the final dataset.

With these summaries, we generate notifications with tailored prompts (see Prompt 1). Subsequently, we rate comprehension progression for each notification, identifying actionable indices and total lengths (see Prompt 2).

\begin{tcolorbox}[colback=red!5!white,colframe=red!40!black,title= Input text: Domain task and missing information]

Given the following environment and control context information:\\
Domain description:\texttt{[Lunar Lander domain simulates a spacecraft attempting to land on the Moon. The agent (lander) is represented as a small, two-legged module navigating a 2D plane...]}\\
State:\texttt{[The state is presented by the lander's position, velocity in both axes, and orientation (angle)...]}\\
Actions:\texttt{[The agent controls the main engine and the two lateral engines...]}\\
Dynamics:\texttt{[There's gravity pulling the lunar lander downwards. Engine thrust counteracts gravity and stabilizes...]}\\
Missing information: \texttt{[The lander does not see the danger zones that are distributed around the trajectory towards the landing pad. \\
There are three danger zones:\\
- Top-left danger zone: above and to the left of the landing pad, near the starting position of the lunar lander.\\
- Overhead blockade: directly above the landing pad.\\
- Right-side obstruction: to the right of the landing pad.]}
\tcbline
Summarize key environmental features:
\begin{itemize}
    \item Extract the critical state variables, action possibilities, and physics constraints that directly influence pilot decisions.
    \item Present each summary point as a brief statement that can serve as the basis for a notification trigger.
\end{itemize}

Integrate Missing Information:
\begin{itemize}
    \item Associate each danger zone with the relevant environmental features and pilot actions.
    \item Rephrase each danger-zone description as a condition that would prompt an urgent notification, emphasizing that the pilot cannot detect these hazards without assistance.
\end{itemize}

\tcbline

Generate your notifications now:
\end{tcolorbox}

First, we describe the environment, task, and any missing information provided as user inputs. We leverage text-based descriptions rather than raw numeric values since our taxonomy primarily involves actionable instructions and context, as is common in LLM-based pipelines. 

\begin{tcolorbox}[colback=green!5!white,colframe=green!40!black,title= Prompt 1: Generate notifications with topic]
    Context: \texttt{[Domain Summary]}
    \tcbline
    Generate \texttt{[Number Samples]} candidate notification(s) instructing the pilot such that the pilot would:
    \begin{itemize}
        \item Each must prompt the pilot to perform the action “\texttt{[Command Type]}.”
        \item Structure each notification so that by the \texttt{[Reaction Time Word]}-th word, the pilot has fully grasped the command.
        \item Use simple, unambiguous phrasing.
        \item Keep each notification under \texttt{[Word Length]} words total.
        \item In at least some of the samples, weave in one piece of the previously missing information to reinforce future decision‐making
    \end{itemize}
    Format: Return each notification as a single line. 
    \tcbline
    Examples of various types of notifications for different instructions:
    \begin{itemize}
        \item ``Pilot, stop now!"
        \item ``Please slow down immediately."
        \item ``There is an obstacle ahead, please go make a shift to the right such that you can avoid it."
        \item ``Please maneuver to the left in order to avoid crossing the flight trajectory of another aircraft."
        \item ``You will be crossing the danger zone in front of you, so descend now."
    \end{itemize}
    Don't number the notifications. Don't provide reasoning or explanations. Only output notifications. 
    \tcbline
    Generate your notifications now:
\end{tcolorbox}

Notifications are iteratively generated until each property reaches a specified threshold. Once surpassed, we perform clustering to remove similar notifications, retaining only the $k$ medoids. If, after 1000 iterations, the threshold remains unmet, we retain all generated notifications, indicating their relative rarity. Detailed prompts, templates, and implementation code are provided in the supplementary materials.

\begin{tcolorbox}[colback=blue!5!white,colframe=blue!40!black,title= Prompt 2: Rating comprehension progression]
    You are given a notification, each an instruction (or possibly multiple instructions) for the pilot. Your task is to analyze each notification word by word and compute how a human's comprehension in understanding what the instruction wants them to do evolves from the first to the last word.\\
    Process:
    \begin{enumerate}
    \item Start at 0\% comprehension before reading any words.
    \item Read the instruction word by word from left to right.
    \item At each word, adjust the comprehension level based on how that word clarifies or obscures the intended action:
    \begin{itemize}
        \item Words that clearly indicate the core action (e.g., ``Reduce '', ``Slow '', ``Descend '') increase comprehension significantly.
        \item Neutral words or filler words that do not add clarity may only slightly increase or keep comprehension stable.
    \end{itemize}
    \item Continue updating the comprehension level word-by-word until the final word.
    \item Return the final results for all notifications as a list of lists of comprehension values (0-100\%) after each word in the sentence. Each sub-list corresponds to one notification. 
    \end{enumerate}
    \tcbline
    In-Context Examples (Guides):\\
    Here are some examples for the comprehension progression for the command type ``Slow down''.
    \begin{itemize}
    \item Instruction: ``Immediate speed reduction needed, danger!''\\
      Words: Immediate $|$ speed $|$ reduction $|$ needed, $|$ danger!\\
      Comprehension after each word: 5\% $|$ 20\% $|$ 70\% $|$ 80\% $|$ 100\%
    \item Instruction: ``Adjust speed, prepare to avoid the zone. ''\\
      Words: Adjust $|$ speed $|$ prepare $|$ to $|$ avoid $|$ the $|$ zone.\\
      Comprehension after each word: 5\% $|$ 50\% $|$ 50\% $|$ 50\% $|$ 70\% $|$ 70\% $|$ 70\%
    \end{itemize}
\end{tcolorbox}

\begin{tcolorbox}[colback=blue!5!white,colframe=blue!40!black,title= Prompt 2: Rating comprehension progression (cont.)]
    ......
    
    \tcbline
    Your Task:\\
    Apply the same reasoning above to the given notification below for the instruction type: \texttt{[Command Type]}. Start from 0\% comprehension before the first word, and after reading each word, estimate the new comprehension level for instruction type: \texttt{[Command Type]}. Please output the comprehension after each word for each of the following instructions, which is intended to instruct the receiver to ``\texttt{[Command Type]}''.
    \tcbline
    \texttt{[Notification]}
    \tcbline
    Return the final result in this format:\\
    
    Instruction: sentence with n words\\
    Words: word 1 $|$ word 2 $|$ word 3 $|$ word 4 $|$ ... $|$ word n-1 $|$ word n.\\
    Comprehension after each word: number\% $|$ number\% $|$ number\% $|$ number\% $|$ ... $|$ number\% $|$ number\%\\
    
    Do not include any explanations or commentary.
    Only output the data structure.
    
\end{tcolorbox}

\section{Broader impacts}
This work addresses a critical yet underexplored challenge in human-robot teaming: the timing of utterances and the variability in human comprehension and reaction must be considered when providing real-time assistive language. By formalizing the trade-off between timeliness and informativeness, we aim to enable more effective, responsive, and human-aligned assistance in time-critical tasks.

However, our framework raises several important considerations. First, human comprehension and reaction times vary widely across individuals and contexts. A model trained on a specific user population or within a simulated reinforcement learning loop may not generalize to unseen users. For example, as demonstrated in our robustness experiments, our system does not generalize well beyond a certain time delay. Without rigorous validation and adaptation, deploying assistive agents in real-world domains like driving or piloting could be risky, as guidance may be mistimed or misaligned with user expectations. Future work should prioritize personalization and adaptability to ensure agents are appropriately matched to individual human partners.

Second, our approach relies on LLMs to simulate human comprehension and generate candidate notifications. While LLMs provide scalable surrogates for prototyping, they carry inherent biases tied to their training data or alignment. Our work uses Llama-3, and if Llama-3 systematically misjudges when humans comprehend partial utterances, our dataset of notifications could end up confusing, misleading, or even unsafe. This risk is particularly salient for underrepresented groups whose cognitive or linguistic patterns may not be well reflected in LLM pretraining data. Research on LLM alignment and personalization could help mitigate this by generating more representative or personalized datasets.

Finally, prior work in human-robot teaming has shown that humans can become overly reliant on AI partners, accepting their advice even if it leads to negative outcomes. If the assistive notifier issues incorrect or poorly timed advice, and the human fails to override it, the result could be a critical failure. While this concern is not unique to our system, it highlights the need for better calibration for assistive AI, and for future research on the mitigation of over-reliance (such as through explainable AI, interpretability, interface design, or better trust calibration strategies).

\end{document}